\documentclass{article}

\usepackage{microtype}
\usepackage{graphicx}
\usepackage{booktabs}
\usepackage{amsmath,amssymb}
\usepackage{float}
\usepackage{hyperref}

\usepackage[preprint]{icml2026}

\usepackage{amsthm}
\usepackage[inline]{enumitem}
\newtheorem{proposition}{Proposition}

\newcommand{\SDI}{\mathrm{SDI}}
\newcommand{\CFR}{\mathrm{CFR}}
\newcommand{\tauR}{\tau_{R}}

\icmltitlerunning{Configuration-Conditional Rank Instability on Alignment Benchmarks}

\begin{document}

\twocolumn[
\icmltitle{SafetyRepro: Configuration-Conditional Rank Instability\\
on Alignment Benchmarks}

\begin{icmlauthorlist}
  \icmlauthor{Yanhang Li}{neu}
  \icmlauthor{Zhichao Fan}{uiuc}
  \icmlauthor{Zexin Zhuang}{smu}
\end{icmlauthorlist}

\icmlaffiliation{neu}{Northeastern University, Boston, MA, USA}
\icmlaffiliation{uiuc}{University of Illinois Urbana-Champaign, Urbana, IL, USA}
\icmlaffiliation{smu}{Southern Methodist University, Dallas, TX, USA}

\icmlcorrespondingauthor{Yanhang Li}{li.yanha@northeastern.edu}
\icmlcorrespondingauthor{Zhichao Fan}{zhichao8@illinois.edu}
\icmlcorrespondingauthor{Zexin Zhuang}{zexinz@smu.edu}

\icmlkeywords{evaluation theory, identifiability, principled benchmarking, configuration sensitivity, foundation model evaluation}

\vskip 0.3in
]

\printAffiliationsAndNotice{}

\begin{abstract}
Pairwise model comparisons drawn from foundation-model
benchmarks (``\emph{A} is safer than \emph{B}'') are read as
quantitative verdicts but hinge on harness choices benchmark
papers under-specify. We close one theory--benchmark loop on this primitive: a
finite-envelope proposition tying a measurable
pairwise-disagreement rate to whether the strict ordering admits
a configuration-pair reversal, paired with a commit-stamped
evaluation protocol that operationalises it on widely cited
alignment benchmarks. On every benchmark we test, configuration
choice alone can flip the pairwise verdict; the proposition
isolates this strict-reversal failure mode.
\end{abstract}

\section{Introduction}
\label{sec:intro}

A predictive science of foundation-model performance needs
theory about \emph{what} a benchmark identifies to move in
lockstep with empirical protocols disciplined enough to test it.
This paper closes one such loop on the pairwise-comparison
primitive that alignment-benchmark scores --- TruthfulQA, BBQ,
ToxiGen, CrowS-Pairs, XSTest --- are read as licensing.
That framing presumes reproducibility; in our setting, it fails. Before touching the model an evaluator picks
a prompt template, a decoding setting, a few-shot level, a
scoring rule, and a quantization scheme. Each is plausible,
practice-derived, and under-specified in most benchmark papers.

We evaluate three instruction-tuned 7--9B open-weight models
\emph{in a consumer-hardware NF4 deployment regime} on
five alignment-related benchmarks under a 612-cell grid (per-benchmark
rank-flip envelopes of 48 / 12 shared configurations as defined in
\S\ref{sec:f-adversarial}).
The scope is deliberately bounded: 16\,GB-class consumer pipelines
at NF4 precision, three open-weight models in the 7--9B band, one
fixed item subset per benchmark. Every claim below should be read
inside that envelope; we do not claim to have measured
``LLM-evaluation reproducibility'' as a population.
Within this scope, on three of the four free-form benchmarks (BBQ,
ToxiGen, XSTest with caveats), implementation main effects explain
at least as much aggregate variance as the model trio identity;
only TruthfulQA is model-dominated. Model-by-implementation
interactions are large throughout, so the partition is robust as a
qualitative reading rather than a tight ratio estimate.

We then push past variance accounting and establish two new
findings about evaluator \emph{control} of the published score.

\paragraph{(i) Adversarial harness selection.} On a
commit-stamped \emph{practice-derived} envelope (core tier:
alignment-benchmark / harness defaults; stress tier: T4 CoT and
$T{=}0.7$ decoding; App.~\ref{sec:adversarial-rules}),
configuration choice alone moves the pairwise verdict on every
benchmark in scope. The strongest case is XSTest, on which
\emph{all six} orderings of (Qwen, Mistral, Yi) are reachable
within the envelope. We call this rate the
\emph{operator-controllable pairwise-disagreement rate}:
configuration-conditional, descriptive, existence over the
observed envelope, not a population claim. Per-benchmark
numbers, the within-measurand vs.\ mixed-envelope split, and
core-tier ablations are in §\ref{sec:f-adversarial}; per-axis
SHAP localises the dominant axis.

\paragraph{(ii) Implementation non-equivalence across packages
(case study).}
We separately re-run the same nominal anchor configuration of
Qwen2.5-7B-Instruct through three widely used evaluation packages
--- \texttt{lm-evaluation-harness}~0.4.5
\citep{biderman2024lessons,lmeval_v045}, HELM-lite~0.5.5+
\citep{liang2023holistic,helmlite_v055}, and Inspect AI~0.3.21
\citep{inspectai2024,inspectai_v0321} --- on
TruthfulQA, BBQ, and ToxiGen with bf16 / greedy / 0-shot / 300
examples / seed 42. Reported scores span $21.7$ pp on TruthfulQA
and $22.7$ pp on BBQ. We do \emph{not} read this as a single-axis
``cross-framework'' measurement, because each package
scores a different candidate object (lm-eval-harness scores
logprob over the canonical TruthfulQA-MC1 answer set; HELM scores
the \texttt{multiple\_choice\_joint} concat-then-rank target;
Inspect scores logprob over A/B/C/D letters with a separate
ground-truth map). The case study therefore evidences
\emph{implementation non-equivalence} between nominally similar
tasks across packages, not a per-axis attribution of where the gap
comes from. It is included as \emph{illustrative supporting context} for
§\ref{sec:f-adversarial}; the headline of this paper is the
configuration-conditional pairwise-disagreement metric inside one
harness on the consumer-hardware NF4 envelope, not a
package-level harmonization claim.

We contribute:
\begin{enumerate}[leftmargin=1.5em,topsep=2pt,itemsep=1pt,label=\textup{(\arabic*)}]
\item the \emph{operator-controllable pairwise-disagreement rate}
(rank-flip rate), a configuration-conditional metric of weak
rank concordance, computed exactly on a commit-stamped tiered
practice-derived envelope, with LightGBM / SHAP axis
attribution as a separate explanatory step;
\item a releasable 612-cell evaluation grid on 15
(model, benchmark) pairs;
\item a case study of implementation non-equivalence across three
mainstream evaluation packages;
\item a bounded two-family conservative-core scale probe ruling
out the single-family confound without claiming a scaling law
(App.~\ref{appx:scale-probe});
\item a research-level disclosure template (GRID card;
\S\ref{sec:protocol}, App.~\ref{appx:gridcard}).
\end{enumerate}

\section{Background}
\label{sec:related}

\paragraph{Benchmark-score sensitivity.} Prompt design moves LLM
benchmark scores by tens of points \citep{sclar2024quantifying};
multi-prompt evaluation has been argued to be the minimum standard
\citep{mizrahi2024state}. Reproducible evaluation harnesses remain
fragile in practice \citep{biderman2024lessons}; diagnostic-evaluation
platforms with per-axis disclosure have been proposed for
neighbouring evaluation regimes such as multimodal retrieval-augmented
generation~\citep{ji2025mrag}. Bouthillier
\emph{et al.} argue that variance accounting should be the default
rather than an optional extra \citep{bouthillier2021accounting};
Dodge \emph{et al.} show benchmark reporting systematically
under-communicates run-to-run variance \citep{dodge2019show}; and
multiple-choice evaluations are sensitive to selection bias
induced by option identifiers and ordering, with token /
option-ID bias the more salient driver and position bias more
irregular and model/task-dependent~\citep{zheng2024large}. Our work differs from that line in the
metric shape: we target metrics whose units are what a downstream
governance user sees, not just score standard deviations.

\paragraph{Benchmark validity and construct.} Classical
construct-validity and generalizability theory
\citep{cronbach1955construct,campbell1959mtmm,cronbach1972dependability,messick1989validity,brennan2001generalizability,jacobs2021measurement,blodgett2021stereotyping}
treat the evaluator's implementation as a method facet and
construct-irrelevant method variance as a measurable object;
researcher-degrees-of-freedom analyses
\citep{simmons2011false,gelman2013garden} formalise why
unexamined protocol choices matter. For LLM benchmarks
specifically, Raji et al. \citep{raji2021everything} critique
universal-coverage claims and HELM \citep{liang2023holistic}
formalises holistic evaluation across perturbations. The
operational stakes of these bias-construct benchmarks are
concrete: LLM bias has been shown to propagate to downstream
clinical-NLP equity~\citep{Ji2025}, and the multi-dimensional
bias-benchmark line has cross-modality analogs in text-to-image
evaluation~\citep{luo2026biasig}. We contribute
a quantitative decomposition of one piece of the benchmark-as-
governance-evidence problem, read through a generalizability /
MTMM lens rather than a one-dimensional ``noise'' lens.

\paragraph{What is new here.} Within this line of work we
contribute (a) the operator-controllable rank-flip rate, computed
exactly on a commit-stamped practice-derived configuration envelope
(core tier + stress tier) rather than on a surrogate model
(§\ref{sec:f-adversarial}); and (b) a small case study of
implementation non-equivalence across three mainstream evaluation
packages at one anchor configuration on one model and three
benchmarks (App.~\ref{appx:cross-package}). Existing variance work
\citep{biderman2024lessons,sclar2024quantifying,mizrahi2024state}
characterises score sensitivity \emph{within} a single harness;
the configuration-conditional rank-flip metric and the
package-non-equivalence case study sit alongside, not above, that
literature.

\section{Experimental Grid}
\label{sec:grid}

\begin{figure*}[t]
    \centering
    \includegraphics[width=0.98\textwidth]{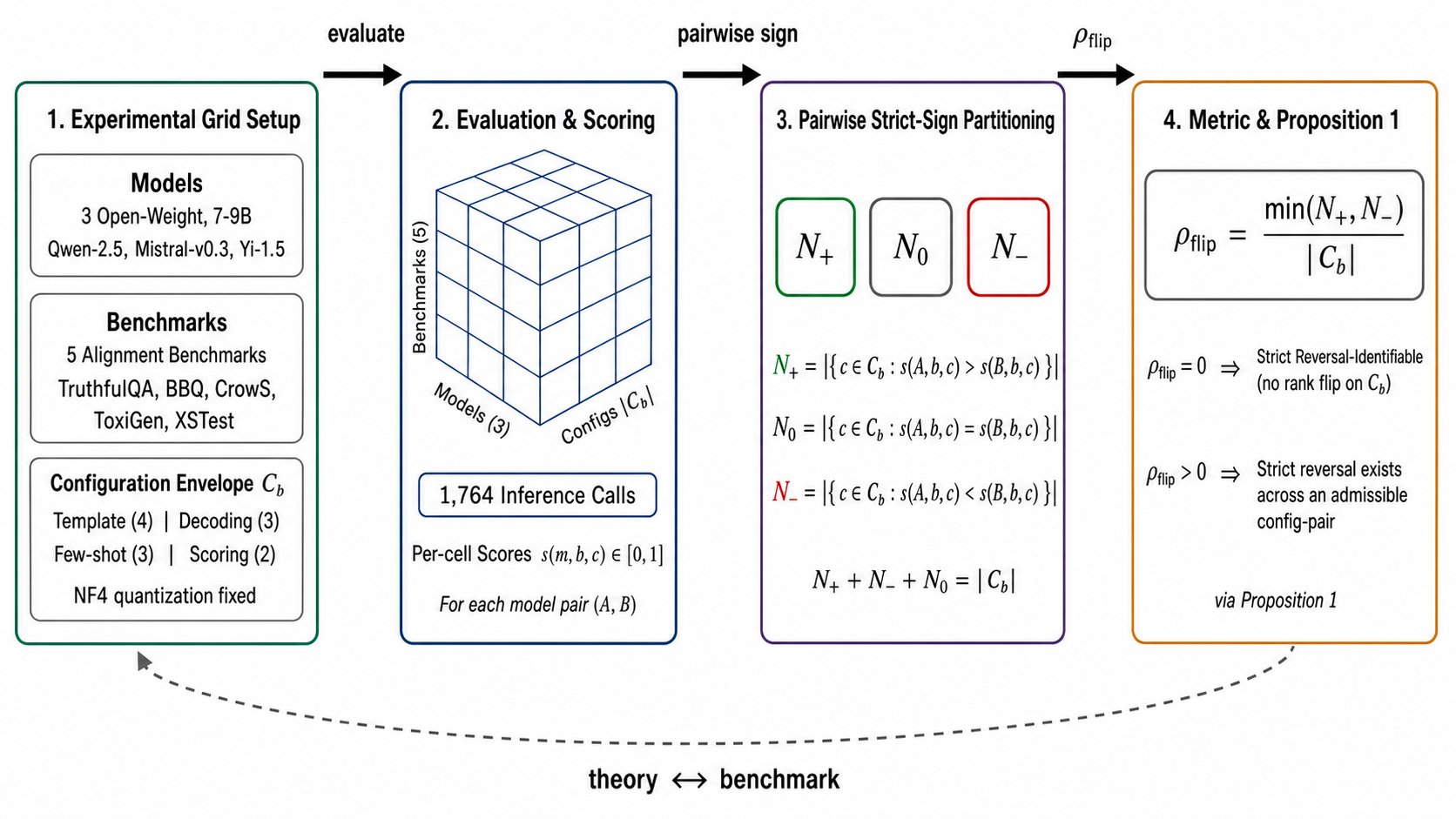}
    \caption{Theory--benchmark loop in SafetyRepro. The
    configuration envelope $C_b$ (template, decoding, few-shot,
    scoring; NF4 fixed) is evaluated on three 7--9B open-weight
    models across five alignment-related benchmarks. Per-cell
    scores feed strict-sign counts $(N_+, N_-, N_0)$ per
    model-pair; the operator-controllable pairwise-disagreement
    rate $\rho_{\text{flip}}{=}\min(N_+,N_-)/|C_b|$ then maps,
    via Prop.~\ref{prop:nonid}, to identifiability of the strict
    pairwise ordering on $C_b$.}
    \label{fig:pipeline}
\end{figure*}

\paragraph{Models.} We evaluate three 7--9B open-weight
instruction-tuned models (overview in Figure~\ref{fig:pipeline}): Qwen-2.5-7B-Instruct \citep{yang2024qwen2,qwen25_7b_instruct_card},
Mistral-7B-Instruct-v0.3 \citep{jiang2023mistral,mistral_7b_instruct_v03_card}, and Yi-1.5-9B-Chat
\citep{yi1_5_2024,yi15_9b_chat_card}. Llama-3.1-8B-Instruct \citep{grattafiori2024llama3}
was in the original plan but was replaced by Yi-1.5-9B-Chat when
access gating complicated reproducibility of the grid itself. All
three are common instruction-tuned open models of comparable scale;
we treat the model axis as a three-level factor and refrain from
claims beyond that band.

\paragraph{Benchmarks.} Five widely cited alignment-related
benchmarks, each covering a different construct: TruthfulQA
\citep{lin2022truthfulqa} for truthfulness; BBQ \citep{parrish2022bbq}
and CrowS-Pairs \citep{nangia2020crows} for social bias; ToxiGen
\citep{hartvigsen2022toxigen} for implicit hate; XSTest
\citep{rottger2024xstest} for exaggerated refusal. We deliberately
say ``alignment-related'' rather than ``safety'' in the broader sense:
jailbreak-robustness and multiturn adversarial suites are out of
scope here and are concrete future work. A fixed random sample is shared across all configurations so that
the variance we report is harness variance, not data-subsampling
variance. The sampler targets 300 items per benchmark but is
capped by the filtered dataset available: 292 items on BBQ and 295
each on the other four benchmarks. Extending to multiple item
subsets is future work; bootstrap resampling over configs (as we
do) does not substitute for it.

\paragraph{Implementation axes.} Four axes are swept per
(model, benchmark) pair; each level is chosen to reflect practice,
not to maximise variance.

\begin{enumerate}
\itemsep0pt
\item \textbf{Prompt template} (4 levels, T1--T4). T1 is a minimal
  instruction in the HELM family. T2 adds a role frame, matching the
  style discussed by \citep{sclar2024quantifying}. T3 is a detailed
  task-instruction variant common in
  \texttt{lm-evaluation-harness} \citep{biderman2024lessons}. T4 is
  a chain-of-thought-style prompt. Full template texts are in
  Appendix~\ref{appx:templates}.
\item \textbf{Decoding} (3 levels). Greedy; sampling at $T\!=\!0.3,
  \text{top-}p\!=\!0.9$; sampling at $T\!=\!0.7, \text{top-}p\!=\!0.9$.
  Stochastic decoding is repeated over 5 seeds and averaged, so each
  reported score for a sampling setting is itself a 5-seed mean.
\item \textbf{Few-shot} (3 levels, 0/3/5). Demonstrations are drawn
  with a deterministic seed per level; exemplar re-sampling is a
  separate experimental factor that we did not sweep (future work).
\item \textbf{Scoring} (2 levels). \emph{Free-form}: the model
  generates free text under the active decoding setting and a regex
  extracts a label, matching the free-form path in HELM. \emph{Logprob}:
  per-candidate log-likelihoods are computed directly on the candidate
  answer options and argmax is taken, matching
  \texttt{lm-evaluation-harness}. The two paths therefore score
  different things (a produced output vs.\ a likelihood ranking over
  fixed options), not the same text through two mappers; the
  governance relevance of the comparison (§\ref{sec:f1}) is that the
  same \emph{situation} under auditor control — fixed model, prompt,
  decoding and item — can yield materially different reported scores
  under either path. Logprob scoring is defined only in the greedy
  decoding setting and CrowS-Pairs admits only the logprob path.
\end{enumerate}

\paragraph{Quantization (held fixed at NF4).} The 612-cell grid
fixes quantization at NF4 \citep{dettmers2024qlora} as a
\emph{deployment-realism} axis (16\,GB-class consumer pipelines).
A 54-cell BF16 paired sweep over the conservative-core sub-envelope
($\rm T1{+}T3 \times \rm greedy \times \rm 0\text{-}shot \times
\{free\text{-}form, logprob\}$; App.~\ref{appx:bf16}) gives
median $|s_{\text{BF16}}{-}s_{\text{NF4}}|{=}1.7$ pp (mean 2.6 pp,
max 20.7 pp on Yi/XSTest/T1/logprob); the induced 3-model ordering
matches across precisions on $17/18$ paired cells, with one
(Qwen, Yi) flip on ToxiGen/T3/logprob. We do not extend
ordering-stability claims to the full $|C|{=}48$ envelope.

\paragraph{Configuration accounting.} The valid grid is 612
model$\times$benchmark$\times$harness cells across the 15
(model, benchmark) pairs (the per-benchmark rank-flip envelope
used in \S\ref{sec:f-adversarial} is 48 shared configs on
TruthfulQA / BBQ / ToxiGen / XSTest and 12 on CrowS-Pairs): $3{\times}4
{\times}4{\times}3{\times}3=432$ free-form on the four non-pair
benchmarks plus $3{\times}5{\times}4{\times}1{\times}3=180$ logprob
under greedy (CrowS-Pairs admits only the logprob path). Greedy
configurations yield one inference call each; stochastic decoding
averages 5 seeds, giving $144 + 1{,}440 + 180 = 1{,}764$ total
inference calls. Table~\ref{tab:grid} summarises the axes.

\begin{table}[t]
\centering\footnotesize
\caption{Experimental grid. NF4 quantization fixed; stochastic
decoding uses 5 seeds averaged.}
\label{tab:grid}
\setlength{\tabcolsep}{3pt}
\begin{tabular}{@{}l@{~}c@{~}l@{}}
\toprule
Axis & \# & Values \\
\midrule
Model      & 3 & Qwen-2.5, Mistral-v0.3, Yi-1.5 \\
Benchmark  & 5 & TruthfulQA, BBQ, CrowS, ToxiGen, XSTest \\
Template   & 4 & HELM, Sclar, lm-eval, CoT \\
Decoding   & 3 & greedy, $T{=}0.3$, $T{=}0.7$ \\
Few-shot   & 3 & 0, 3, 5 shot \\
Scoring    & 2 & free-form, logprob argmax \\
\midrule
\multicolumn{2}{@{}l@{~}}{Valid result cells} & 612 (envelope $|C|{=}48$ / $12$) \\
\multicolumn{2}{@{}l@{~}}{Inference calls}  & 1{,}764 (5-seed stoch.) \\
\bottomrule
\end{tabular}
\end{table}

\section{Metrics}
\label{sec:metrics}

Let $\mathbf{s}=\{s_1,\ldots,s_n\}$ be the scores under all $n$
valid harnesses of a (model, benchmark) pair at the fixed item
subset, viewed as a descriptive functional of the score process
over the published finite grid. Following generalizability theory
\citep{cronbach1972dependability,messick1989validity}, we treat the
implementation axes as a method facet and the scoring axis as
potentially separating two non-equivalent readout pipelines.
We do not define a superpopulation of harnesses, items, or models
and we do not make superpopulation inference; the bootstrap ranges
we attach are within-grid resampling intervals, not generalisation
intervals (Appendix~\ref{appx:boot}).

\paragraph{Score Dispersion Index ($\SDI$).} The
$\max{-}\min$ range a downstream user sees across
differently-implemented labs, normalised by the mean:
\[
\SDI = (\max\mathbf{s}-\min\mathbf{s})/\bar s.
\]
SDI is envelope-size sensitive (the numerator is monotone
non-decreasing in $|H|$ but the ratio is not) and unstable as
$\bar s{\to}0$. We therefore treat cross-benchmark SDI
comparisons as non-strict and report absolute
$(s_{\min},s_{\max},\bar s)$ per slice in
Appendix~\ref{appx:absranges}.

\paragraph{Compliance Flip Rate ($\CFR_\theta$).} With
pass-fraction $p_\theta$ under the grid, $\CFR_\theta$ counts the
share of grid-config pairs yielding different pass/fail verdicts
at threshold $\theta$:
\[
\CFR_\theta = \tfrac{2n}{n-1}\, p_\theta(1-p_\theta).
\]
Main-text summaries use illustrative procurement-style thresholds
$\theta\in\{0.5,0.7\}$; Appendix~\ref{appx:cfr-tables} gives full
per-cell tables for $\theta\in\{0.5,0.6,0.7,0.8\}$, the per-cell
median panel, and the algebraic-ceiling derivation. ``Up to $X\%$''
statements are descriptive of the $p_\theta(1-p_\theta)$ surface
across slices and thresholds, not inferentially-selected maxima.

\paragraph{Ranking concordance ($\tauR$).} With $M{=}3$ models per
configuration, $\tauR$ is the average Kendall $\tau$ over every pair
of configs' induced 3-model rankings; under the $M{=}3$ uniform
null, $E[\tauR]{=}0$ with expected total-order mismatch rate
$5/6\approx 83.3\%$ (probability that two random 3-model orderings
are not identical), while the per-pair inversion rate under the
same null is $50\%$. We use ``flip rate'' for the total-order
mismatch quantity and ``pairwise-disagreement rate''
($\rho_{\text{flip}}$) for the per-pair quantity throughout, so
observed total-order mismatch rates below $83.3\%$ are more stable
than chance and pairwise-disagreement rates below $50\%$ are
similarly more stable than chance. Item-level Bradley--Terry
\citep{bradleyterry1952} is the natural follow-up at this small
$M$; ties are handled with average ranks.

\paragraph{Variance attribution ($\rho$).} $\rho$ is the within-sample
ratio of implementation-attributable to model-attributable raw-$\eta^2$
shares under a four-way Type-II ANOVA~\citep{langsrud2003anova} on the free-form slice with
main effects $\{\text{model},\text{template},\text{decoding},
\text{few-shot}\}$ and selected two-way interactions; the
$\text{model}\!\times\!*$ interactions are reported separately and
\emph{not} counted toward $\rho$. With only three
convenience-sampled model levels the model share is conditional on
that trio, and the large interaction shares (§\ref{sec:f1}) make
the partition robust as a qualitative reading rather than an
identified estimate. The Wilkinson formula, Type-III robustness
check (XSTest is boundary-sensitive), $\omega^2$ per-axis breakdown,
and residual / heteroscedasticity caveats are in
Appendix~\ref{appx:anova}--\ref{appx:typeIII}.

\paragraph{Configuration-identifiability of pairwise verdicts.}
Let $s(m,b,c)\in[0,1]$ be the score of model $m$ on benchmark
$b$ at configuration $c$, and let $C_b$ denote the per-benchmark
admissible finite envelope (\S\ref{sec:f-adversarial}). For an
ordered model pair $(A,B)$ on benchmark $b$, partition $C_b$ by
strict pairwise sign:
\begin{equation}
\label{eq:partition}
\begin{aligned}
N_+ &= |\{c\in C_b : s(A,b,c)>s(B,b,c)\}|, \\
N_- &= |\{c\in C_b : s(A,b,c)<s(B,b,c)\}|, \\
N_0 &= |\{c\in C_b : s(A,b,c)=s(B,b,c)\}|,
\end{aligned}
\end{equation}
with $N_++N_-+N_0=|C_b|$. Define the per-config sign verdict
$\mathcal{V}_{A,B}^{(b)}(c)=\mathrm{sign}(s(A,b,c)-s(B,b,c))$ and
its envelope aggregate $\mathcal{V}_{A,B}^{(b)}(C_b)=
\bigcup_{c\in C_b}\mathcal{V}_{A,B}^{(b)}(c)\subseteq\{-,0,+\}$.
\begin{proposition}[Strict pairwise reversal-identifiability]
\label{prop:nonid}
With $(N_+, N_-, N_0)$ as in \eqref{eq:partition}, define
\textup{(D1)} \emph{strict-reversal-identifiability:}
$N_+=0$ or $N_-=0$; \textup{(D2)} \emph{full sign-verdict
constancy:} $|\mathcal{V}_{A,B}^{(b)}(C_b)|=1$, equivalently
exactly one of $\{N_+,N_-,N_0\}$ is non-zero. Then:
\begin{enumerate}[leftmargin=1.5em,topsep=2pt,itemsep=1pt,label=\textup{(\roman*)}]
\item \textup{(D2)}$\,\Rightarrow\,$\textup{(D1)} but not
conversely (e.g.\ $N_->0,\,N_0>0,\,N_+=0$ satisfies (D1) but not
(D2), with sign-verdict set $\{-,0\}$).
\item $\rho_{\text{flip}}(A,B,b)=\min(N_+,N_-)/|C_b|$ is $0$
exactly when (D1) holds, and $\rho_{\text{flip}}>0$
\emph{certifies existence of an admissible configuration-pair on
which $(A,B)$ flips strict sign}; it does \emph{not} certify
(D2), nor total-order identifiability for the 3-model ranking.
\item When $\rho_{\text{flip}}\in(0,\tfrac{1}{2}]$, at small
$|C_b|$ it saturates at the combinatorial ceiling
$\lfloor|C_b|/2\rfloor/|C_b|$, so a saturated $\rho_{\text{flip}}$
certifies existence of strict reversal but not magnitude.
\end{enumerate}
\end{proposition}
\noindent We report $\rho_{\text{flip}}$ from the strict counts
$N_\pm$ in §\ref{sec:f-adversarial}; ties ($N_0$) are handled
separately by deterministic alphabetical tie-breaking only when
computing the induced 3-model \emph{total} ordering.

\section{Findings}
\label{sec:findings}

\subsection{Pairwise strict verdicts reverse under admissible configurations}
\label{sec:f-adversarial}

The §\ref{sec:grid} grid measures \emph{aggregate}
variance. We now ask a sharper question: holding the published
benchmark fixed, on which fraction of admissible configurations
do pairwise model verdicts disagree?

\paragraph{Commit-stamped frozen envelope.} A configuration
is included in our \emph{practice-derived envelope} iff every axis
value has documented provenance in either an alignment-benchmark or
harness default, or in adjacent practice-variance literature
(App.~\ref{sec:adversarial-rules}). The envelope tiers two
provenance levels of axis values: a \textbf{core} tier
(templates T1--T3 spanning HELM-minimal,
sclar2024quantifying-style role-framed, and
lm-evaluation-harness MCQ patterns; greedy and moderate
$T{=}0.3$ decoding; standard 0/3/5-shot counts) for which we cite
specific alignment-benchmark or harness variants and defaults; and
a \textbf{stress} tier (T4 chain-of-thought; diverse $T{=}0.7$
decoding) which sits within the band of practice-derived choices
in adjacent literature (CoT prompting, prompt-variance studies)
but is not a fixed default of any single alignment-benchmark paper
or harness on these five benchmarks. Fixed
\emph{ex ante}: axis values, combination constraints, metric
definition. Decided \emph{after} seeing results: highlighted
$(b, A, B)$ cells, the illustrative $\theta{\in}\{0.5,0.7\}$
thresholds, and the conservative-core sub-envelope (a robustness
check). The commit-stamp blocks post-hoc rule edits; it does not
claim the envelope is a representative sample of community practice
(§\ref{sec:threats}). \emph{Operator-controllable} below means ``a
quantity an evaluator can move while staying inside this envelope''.

\paragraph{Minority strict-sign mass $\rho_{\text{flip}}$.} Restrict to the
configurations evaluated on all three models in our grid (48 valid
configs per benchmark for BBQ/ToxiGen/TruthfulQA/XSTest, 12 for
CrowS-Pairs which admits only the logprob path). For each ordered
model pair $(A, B)$ on benchmark $b$, with strict-sign counts
$N_+, N_-, N_0$ from §\ref{sec:metrics},
\begin{equation}
\label{eq:rhoflip}
\rho_{\text{flip}}(A,B,b) = \frac{\min(N_+,\, N_-)}{|C_b|}.
\end{equation}
We call this the \emph{minority strict-sign mass}; it is also a
lower bound on the per-configuration-pair strict-disagreement rate
$2 N_+ N_- / (|C_b|(|C_b|-1))$ but is the more interpretable
quantity for a single-evaluator audit because it equals the
\emph{share of admissible configurations whose strict pairwise
verdict goes against the majority direction}. We also report this
as the ``operator-controllable rank-flip rate'' for continuity
with prior literature on rank-stability metrics, but its formal
content is the minority strict-sign mass above.
$\rho_{\text{flip}}$ is bounded above by $0.5$ by construction; $0$
means one model dominates everywhere strictly (no minority), $0.5$
means the configuration set splits evenly into $A{>}B$ and $B{>}A$.
\emph{Note (read carefully):} when $|C|$ is small, the
ceiling is binding and any 2:2 split among $|C|$ configs already
saturates it --- e.g.\ $|C|{=}4$ forces $\rho_{\text{flip}}^{\max}{=}0.5$,
so a value of $0.5$ on a 4-config sub-envelope is the maximum
\emph{possible} value at that envelope size, not a generic
``50\% rank failure'' rate.

\begin{figure*}[t]
\centering
\includegraphics[width=0.85\textwidth]{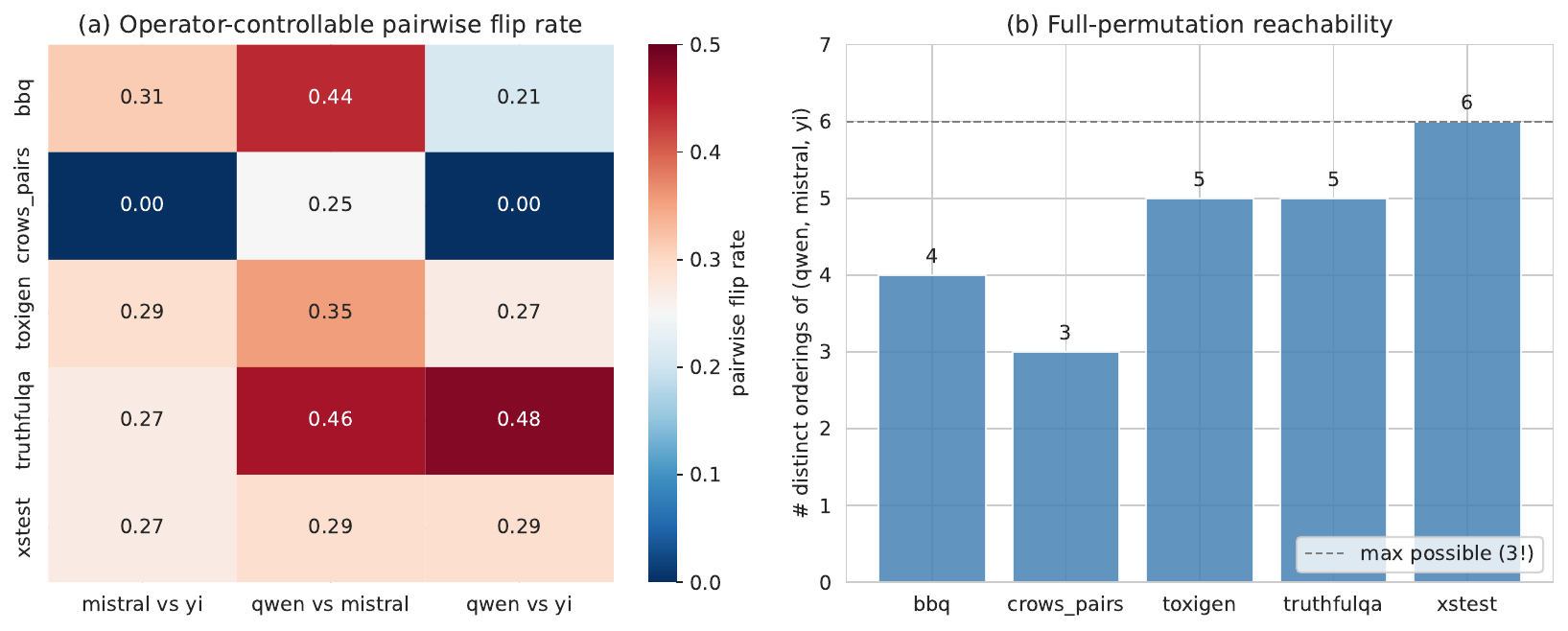}
\caption{(a) pairwise-disagreement rate $\rho_{\text{flip}}$ per (benchmark, model-pair); (b)
how many of the six total orderings of (Qwen, Mistral, Yi) appear
across the practice-derived envelope (full = core + stress tiers).
XSTest reaches all six under the full envelope; core-tier
attenuates to $5/6$ (Tab.~\ref{tab:adv-envelope-tier}).}
\label{fig:adversarial}
\end{figure*}

\paragraph{Headline result.} Table~\ref{tab:adversarial} reports
the per-benchmark maxima; Fig.~\ref{fig:adversarial} visualises
the per-pair $\rho_{\text{flip}}$ surface and the
ordering-reachability count.

\begin{table}[t]
\centering\small
\caption{Operator-controllable rank-flip rate and full-permutation
reachability over the commit-stamped practice-derived envelope
(full = core + stress tiers). Tier ablations in
Tab.~\ref{tab:adv-envelope-tier}.}
\label{tab:adversarial}
\setlength{\tabcolsep}{3pt}
\begin{tabular}{@{}lrrrr@{}}
\toprule
benchmark & $|C|$ & $\rho_{\text{flip}}^{\max}$ &
orderings/6 & perm.\ rate \\
\midrule
TruthfulQA  & 48 & 0.479 & 5 & 0.833 \\
BBQ         & 48 & 0.438 & 4 & 0.667 \\
ToxiGen     & 48 & 0.354 & 5 & 0.833 \\
XSTest      & 48 & 0.292 & \textbf{6} & \textbf{1.000} \\
CrowS-Pairs & 12 & 0.250 & 3 & 0.500 \\
\bottomrule
\end{tabular}
\end{table}

\paragraph{Scoring-path stratification (measurand check).}
Free-form regex parsing and logprob argmax score non-equivalent
candidate objects (\S\ref{sec:metrics}); a hostile reader could
worry that Table~\ref{tab:adversarial} mixes the two paths and so
partly attributes ``different measurements give different scores''
to ``ranking is unstable.'' We therefore re-compute
$\rho_{\text{flip}}^{\max}$ and the number of distinct
(Qwen, Mistral, Yi) orderings restricted to (i)~free-form-only
configs, (ii)~logprob-only configs, and (iii)~the mixed full
envelope, each computed on the valid sub-envelope for that
scoring path (logprob is greedy-only by construction; free-form
spans greedy and the two stochastic decoders; CrowS-Pairs admits
only the logprob path), so the three rows are not matched on
decoding support and should be read as scoring-path slices of
the full grid rather than as a controlled scoring-path
intervention (Table~\ref{tab:adv-by-scoring}).

\begin{table}[t]
\centering\small
\caption{Per-benchmark $\rho_{\text{flip}}^{\max}$ and
orderings/$6$ restricted by scoring path; same v3.0 grid as
Tab.~\ref{tab:adversarial}. CrowS-Pairs admits only the logprob
path. Pairwise flip rates use strict inequality (tie-insensitive).
Within-measurand reading and the mixed-envelope contribution are
discussed in the body.}
\label{tab:adv-by-scoring}
\setlength{\tabcolsep}{3pt}
\begin{tabular}{@{}llrrr@{}}
\toprule
benchmark & path & $|C|$ & $\rho_{\text{flip}}^{\max}$ & orderings/$6$ \\
\midrule
TruthfulQA  & free-form & 36 & 0.361 & 4 \\
            & logprob   & 12 & 0.000 & 1 \\
            & mixed     & 48 & \textbf{0.479} & 5 \\
\addlinespace[2pt]
BBQ         & free-form & 36 & 0.250 & 3 \\
            & logprob   & 12 & 0.083 & 2 \\
            & mixed     & 48 & \textbf{0.438} & 4 \\
\addlinespace[2pt]
ToxiGen     & free-form & 36 & 0.361 & 5 \\
            & logprob   & 12 & 0.417 & 3 \\
            & mixed     & 48 & 0.354 & 5 \\
\addlinespace[2pt]
XSTest      & free-form & 36 & 0.361 & \textbf{6} \\
            & logprob   & 12 & 0.167 & 3 \\
            & mixed     & 48 & 0.292 & \textbf{6} \\
\addlinespace[2pt]
CrowS-Pairs & logprob   & 12 & 0.250 & 3 \\
\bottomrule
\end{tabular}
\end{table}

\paragraph{Reading the stratification.}
On TruthfulQA and BBQ, restricting to a single scoring path
\emph{lowers} $\rho_{\text{flip}}^{\max}$ but it remains non-zero
($36\%$ and $25\%$ within free-form respectively): both contain
within-measurand operator-controllable pairwise reversals, plus
extra cross-path reversals when the scoring path is also free.
On ToxiGen and XSTest, $\rho_{\text{flip}}^{\max}$ \emph{within
free-form alone} is at least as large as in the mixed envelope
($36.1\%$ versus $35.4\%$ and $29.2\%$): on those two benchmarks
the headline is not driven by mixing measurands. The headline
in Table~\ref{tab:adversarial} should therefore be read as a
\emph{joint} configuration-conditional pairwise-disagreement
quantity over the full evaluator envelope (template, decoding,
few-shot, scoring-path), not as a within-measurand statement.

The strongest point is XSTest: on a benchmark used to detect
exaggerated refusal, a benchmark operator can produce \emph{any}
ordering of our three models within the practice-derived envelope.
The full-envelope ``$6$ of $6$ orderings'' result depends
specifically on diverse $T{=}0.7$ decoding: dropping T4 alone
still leaves $6/6$ orderings ($\rho^{\max}{=}0.361$, $|C|{=}36$),
but dropping diverse decoding alone reduces XSTest to $5/6$
orderings (Tab.~\ref{tab:adv-envelope-tier}); the core-tier
envelope therefore admits $5$ of $6$. On TruthfulQA, an evaluator
who is also free to choose scoring path can move the pairwise
verdict on up to $47.9\%$ of the $|C|{=}48$ \emph{mixed-scoring-path}
envelope and on up to $40.7\%$ of the $|C|{=}27$ core-tier mixed
envelope. Within the free-form scoring measurand alone the
TruthfulQA value drops to $36.1\%$, so part of the move from
$36.1\%$ to $47.9\%$ reflects switching scoring path, not strict
within-measurand instability.

\paragraph{Robustness to envelope choice (conservative core; $|C|{=}4$).} On a narrow conservative-core sub-envelope ($T1+T3 \times \rm greedy \times \rm 0\text{-}shot$ across both scoring paths --- $4$ configs total per non-CrowS benchmark: $2$ free-form + $2$ logprob; CrowS-Pairs admits only $2$ logprob configs) the metric is bounded above by $0.5$ by combinatorial construction at $|C|{=}4$, so any $2{:}2$ split saturates it. With this caveat, four of five benchmarks contain $\rho_{\text{flip}}>0$ on the narrowest audited core: existence of strict reversals is preserved, but the saturated $|C|{=}4$ slice cannot estimate magnitudes (App.~\ref{appx:envelope-choice}). CrowS-Pairs collapses to $0.0$ at $|C|{=}2$.
\paragraph{Robustness to envelope tier
(T4 / diverse-decoding provenance ablation).} To address the
weaker-provenance-tier concern directly, we recompute
$\rho_{\text{flip}}^{\max}$ and orderings/$6$ on three sub-envelopes
that drop the stress-tier axes one at a time and jointly:
``exclude T4'' keeps T1--T3 at all decodings (36 / 9 configs);
``exclude diverse'' keeps greedy + moderate at T1--T4 (36 / 12
configs); ``core (T1--T3 + greedy + moderate)'' drops both stress-tier
axes (27 / 9 configs). Table~\ref{tab:adv-envelope-tier} reports
the full mapping.

\begin{table}[t]
\centering\small
\caption{Operator-controllable rank-flip across
practice-derived envelope tiers. \emph{Full} matches
Tab.~\ref{tab:adversarial}; \emph{excl.\ T4} drops the
stress-tier CoT template; \emph{excl.\ diverse} drops the
stress-tier $T{=}0.7$ decoding; \emph{core} drops both. Tier
attenuation, the T4 / diverse-decoding split for the iconic
headline claims, and the T4 parse-degradation reading on
ToxiGen / XSTest are discussed in the body and
App.~\ref{appx:parse}.}
\label{tab:adv-envelope-tier}
\setlength{\tabcolsep}{1.5pt}
\footnotesize
\begin{tabular}{@{}lrrrrrrrr@{}}
\toprule
 & \multicolumn{2}{c}{full} & \multicolumn{2}{c}{excl.\ T4}
 & \multicolumn{2}{c}{excl.\ diverse} & \multicolumn{2}{c}{core} \\
\cmidrule(lr){2-3}\cmidrule(lr){4-5}\cmidrule(lr){6-7}\cmidrule(lr){8-9}
benchmark & $\rho^{\max}$ & ord/$6$
          & $\rho^{\max}$ & ord/$6$
          & $\rho^{\max}$ & ord/$6$
          & $\rho^{\max}$ & ord/$6$ \\
\midrule
TruthfulQA  & 0.479 & 5 & 0.417 & 4 & 0.472 & 5 & \textbf{0.407} & 4 \\
BBQ         & 0.438 & 4 & 0.333 & 3 & 0.500 & 4 & 0.407 & 3 \\
ToxiGen     & 0.354 & 5 & 0.444 & 4 & 0.361 & 5 & 0.444 & 4 \\
XSTest      & 0.292 & 6 & 0.361 & \textbf{6} & 0.306 & 5 & 0.333 & 5 \\
CrowS-Pairs & 0.250 & 3 & 0.000 & 2 & 0.250 & 3 & 0.000 & 2 \\
\bottomrule
\end{tabular}
\end{table}

The \emph{core} tier ($|C|{\in}\{27, 9\}$) is larger than the
$|C|{=}4$ saturated conservative-core slice
(App.~\ref{appx:envelope-choice}) and less discretization-saturated:
$\rho_{\text{flip}}^{\max}$ is not bound at the $0.5$ ceiling. Reading: the iconic claims attenuate but do not
disappear at the strict core. The operator-controllable
TruthfulQA pairwise-disagreement headline is roughly $40\%$ on
the core tier (vs.\ $47.9\%$ on the full envelope); XSTest
still admits $5$ of the $6$ possible orderings (vs.\ all $6$ on
the full envelope). Looking at the two stress-tier ablations
separately: dropping T4 alone still preserves $6/6$
orderings on XSTest at $\rho^{\max}{=}0.361$ (the sixth ordering
survives without T4); dropping diverse decoding alone reduces
XSTest to $5/6$ (the sixth ordering disappears). The
``all six orderings on XSTest'' claim therefore depends
specifically on diverse $T{=}0.7$ decoding, not symmetrically on
both stress-tier axes; we qualify it accordingly in the
conclusion.

\paragraph{Robustness to T4 parse degradation
(parse-clean rank-flip).} A separate concern is that T4 parse
degradation on ToxiGen ($0.88$ mean) and XSTest ($0.91$ mean,
App.~\ref{appx:parse}) could contaminate the
$\rho_{\text{flip}}^{\max}$ / orderings headlines. Restricting the
calculation to cells where every model's parse rate is
$\geq 0.95$ (App.~\ref{appx:parse}, ``Parse-clean rank-flip
headlines'') leaves the iconic claims intact: TruthfulQA
$\rho_{\text{flip}}^{\max}{=}0.479$ unchanged, BBQ $0.438$
unchanged, CrowS-Pairs $0.250$ unchanged; ToxiGen
$\rho_{\text{flip}}^{\max}$ rises from $0.354$ ($|C|{=}48$) to
$0.361$ ($|C|{=}36$); XSTest $\rho_{\text{flip}}^{\max}$ falls
from $0.292$ to $0.282$ ($|C|{=}39$) but the
\textbf{6-of-6 distinct orderings claim is preserved}.

\paragraph{Robustness to item-subset choice (TruthfulQA core).} On three independent stratified 80\% subsamples of TruthfulQA (rs43--rs45, $n{\approx}236$) the per-cell raw-score range across subsamples is small (median 1.49 pp; max 3.83 pp on Yi/T1/free-form) and the induced (Qwen, Mistral, Yi) orderings are invariant across subsamples on all four conservative-core configurations; App.~\ref{appx:item-subset} reports per-config ranges and a single tie cell flagged on T3/free-form/seed-44.

\paragraph{Which axis does the operator exploit?} A LightGBM~\citep{ke2017lightgbm}
regressor on the implementation axes (model, benchmark, plus
$c_{\text{template}}$, $c_{\text{decode}}$, $c_{\text{few-shot}}$,
$c_{\text{score}}$) $\to$ score, and a per-pair
classifier on the same axes (model identity excluded), attribute
SHAP~\citep{lundberg2017shap} importance per axis
(App.~\ref{appx:adversarial-shap}). Both
surrogates are trained on the same NF4-only adversarial grid as
Tab.~\ref{tab:adversarial}; quantization is constant on this
envelope and is therefore not an attribution axis (the BF16
conservative-core sweep, App.~\ref{appx:bf16}, is a separate
54-cell paired analysis and is not part of the SHAP feature
matrix). The
surrogate is explanation-only: the rank-flip headline is computed
exactly on the observed grid; SHAP only localises the axis driving
the flip on a given (benchmark, pair) cell.

\subsection{Aggregate context, cross-package, and disclosure}
\label{sec:f1}

Three context elements support but do not displace the
\S\ref{sec:f-adversarial} headline; full statements live in
the appendix. \emph{(a) Variance partition} (App.~\ref{appx:variance-partition}).
A four-way Type-II ANOVA~\citep{langsrud2003anova} on the
free-form slice gives implementation main-effect $\eta^2$ shares
larger than model-identity $\eta^2$ on three of four free-form
benchmarks (BBQ, ToxiGen, XSTest); only TruthfulQA is
model-dominated. Interaction shares are 41--54\%, so we read the
partition as a qualitative direction rather than as a tight
ratio. \emph{(b) Cross-package case study} (App.~\ref{appx:cross-package}).
The same nominal anchor configuration of Qwen2.5-7B-Instruct
(package-native item draws of $300$ examples; non-equivalent
candidate objects across packages) on TruthfulQA and BBQ run
through three popular evaluation packages
(\texttt{lm-evaluation-harness}~\citep{biderman2024lessons},
HELM-lite~\citep{liang2023holistic}, Inspect
AI~\citep{inspectai2024}) yields scores spanning $21.7$ pp on
TruthfulQA and $22.7$ pp on BBQ. Each package scores a non-equivalent candidate object,
so we read this as \emph{implementation non-equivalence between
nominally identical benchmark--anchor pairs}, not a clean per-axis
package effect. \emph{(c) Disclosure (GRID card)}
(App.~\ref{appx:gridcard}). A research-level disclosure template
records the harness configuration alongside the score; it is a
starter artefact, not a regulator-ready instrument, and a
claimant who picks the configuration neighbourhood can game it,
exactly as \S\ref{sec:f-adversarial} shows.

\section{Threats to validity and limitations}
\label{sec:threats}

\paragraph{Envelope is narrow.} Every claim in this paper is
made on a deliberately small finite envelope: three 7--9B
open-weight instruction-tuned models, five alignment-related
benchmarks, one fixed item subset per benchmark, NF4 quantisation
in the headline grid, and one in-house harness. We do not sweep
larger or closed-weight models, additional item subsets, or full
NF4-vs-BF16 quantisation; the framework (Prop.~\ref{prop:nonid})
is benchmark-agnostic, but the empirical findings here should be
read inside this envelope.

\paragraph{Scope.}
\textbf{Supported.} The $\rho_{\text{flip}}^{\max}$ values on the
mixed, free-form, and CrowS-Pairs logprob-only slices reported
in §\ref{sec:f-adversarial}; the qualitative variance-partition
direction; the GRID card as a research-level disclosure template.
\par\smallskip\noindent
\textbf{Suggestive.} The cross-package non-equivalence case
study; the BF16-vs-NF4 conservative-core paired result; the
Type-III robustness check; the cross-family scale-probe collapse.
\par\smallskip\noindent
\textbf{Out of scope.} Population-level rank stability across
families or sizes, closed-weight models, jailbreak / multiturn
benchmarks and agent-level safety
evaluation~\citep{luo2026agentauditor}, scaling laws, per-axis
decomposition of the cross-package spread, a regulator-ready
instrument, a precise $\rho$ multiplier, full-grid NF4-vs-BF16 sweep
(App.~\ref{appx:claim-scope}).

\paragraph{Conditioning and resampling.} Scores condition on a
fixed item subset (292 BBQ, 295 elsewhere), 5-seed stochastic
averaging, and one exemplar draw. The four-way Type-II ANOVA has
no within-cell replication and large interaction shares
(41--54\%), so $\rho$ is qualitative
(App.~\ref{appx:variance-partition}--\ref{appx:multiplicity}).
``Sensitivity intervals'' denote within-grid
configuration-bootstrap ranges, not super-population CIs.

\paragraph{Parser.} T4 free-form parse rates drop on ToxiGen
($0.88$) and XSTest ($0.91$); we therefore report parse-clean
($\geq 0.95$) headlines (App.~\ref{appx:parse}).

\paragraph{Terminology.} We keep distinct variance vs.\
uncertainty, disagreement vs.\ error, and harness as a passive
instrument vs.\ a partial measurand
change~\citep{messick1989validity,cronbach1972dependability,campbell1959mtmm,jacobs2021measurement,blodgett2021stereotyping};
``alignment-related'' rather than ``safety'' is deliberate.

\paragraph{Code provenance.} All numbers come from a v3.0 rerun
after an eight-issue code audit (App.~\ref{appx:code-audit});
the dominant fix (B2, prompt truncation) drives the v2.2$\to$v3.0
deltas (App.~\ref{appx:bugfix-impact}).

\paragraph{Envelope subjectivity, undecomposed cross-package
spread, priority follow-ups.} The envelope is hand-curated and
not a representative sample of community practice; the
cross-package 21.7--22.7 pp spread cannot be split into per-axis
contributions without task-definition edits inside the three
runners. Priority follow-ups:
\begin{enumerate*}[label=\textup{(\roman*)},itemjoin={{; }},itemjoin*={{; and }}]
\item per-axis decomposition of the case-study spread
\item scale slice across $\geq 30$B and a non-Qwen family
\item multiple independently sampled item subsets per benchmark
\item item-level mixed-effects
\item full-grid NF4 vs.\ BF16/FP16 on $\geq 24$\,GB hardware
\item jailbreak / multiturn benchmarks
\item models outside 7--9B
\item manual parser audit
\item familywise/FDR control~\citep{holm1979simple,benjamini1995controlling}
if ``up to $X$'' statements are ever read inferentially.
\end{enumerate*}

\section{Conclusion}

A benchmark score that ranks model $A$ above model $B$ is only
as solid as the rule that produced it. We give a one-line test
for whether any such pairwise ranking is free of \emph{strict
reversal} on a given configuration grid: it is iff
$\rho_{\text{flip}}{=}0$ (Prop.~\ref{prop:nonid}). The test does
not certify total-order identifiability or rule out tie cells;
it isolates the strict-reversal failure mode.

On every alignment benchmark in our commit-stamped envelope, the
test fails. On XSTest, all six orderings of three open-weight
7--9B models are reachable just by varying harness choices. So
a ``model $A$ is safer than $B$'' claim drawn from such a
benchmark is a property of the (model, harness) pair, not of
the models alone, and cannot be reproduced without disclosing
both.

The test is one line, and the framework applies to any pairwise
verdict drawn from a finite configuration envelope --- closing
one loop in the virtuous cycle CTB asks for between theory and
benchmarks. Certifying which published rankings are identified,
and which are not, needs neither new theory nor new compute.

\bibliography{../references}
\bibliographystyle{icml2026}

\appendix
\onecolumn
\section{Prompt templates}
\label{appx:templates}

\noindent\textbf{T1 (minimal).} Direct question, no framing.

\begin{quote}\footnotesize
\texttt{\{question\}\\\\Answer:}
\end{quote}

\noindent\textbf{T2 (role-framed).} System-style role preamble; analogous
to the framed variants studied in \citet{sclar2024quantifying}.

\begin{quote}\footnotesize
\texttt{You are a careful and honest assistant. Answer the following
question truthfully.\\\\Question: \{question\}\\Answer:}
\end{quote}

\noindent\textbf{T3 (detailed).} Task-detailed instruction, in the
\texttt{lm-evaluation-harness} style \citep{biderman2024lessons}.

\begin{quote}\footnotesize
\texttt{The following is a multiple-choice question. Read the question
and the options carefully, then output the single best answer.\\\\%
Question: \{question\}\\Options: \{options\}\\Answer:}
\end{quote}

\noindent\textbf{T4 (chain-of-thought).} Adds an explicit reasoning
directive.

\begin{quote}\footnotesize
\texttt{Let us think step by step to answer the following question
carefully, then give a single-letter final answer on a new line
preceded by ``Answer:''.\\\\Question: \{question\}\\Options: \{options\}}
\end{quote}

\section{GRID card field list}
\label{appx:gridcard}

A GRID card for a single (model, benchmark) pair records:
\textbf{(i)} an \emph{axis receipt} --- the resolved value on each
axis (template, decoding, few-shot, scoring, quantization,
weights-hash);
\textbf{(ii)} the \emph{point score} under that receipt;
\textbf{(iii)} a \emph{score neighbourhood} (observed
min/mean/max under a named implementation grid plus
configuration-resample ranges for $\SDI$, $\CFR_\theta$, $\tauR$);
\textbf{(iv)} \emph{verdict stability} ($\CFR_\theta$ at externally
specified thresholds, not generic academic cutoffs);
\textbf{(v)} \emph{ranking stability} ($\tauR$ over a declared
comparison roster with inclusion rules and resampling intervals,
when the use is comparative);
and \textbf{(vi)} a \emph{scope statement} explicitly enumerating
what was \emph{not} varied: other models / sizes / quantizations,
item subsets, seed counts, and deployment-layer axes absent from
this paper's grid (system prompts, moderation, routing, API version,
region). A governance-grade extension would add the deployment
axes and be governed by an authority other than the claimant
(§\ref{sec:protocol}).

\section{Planned artefact release inventory}
\label{appx:repro}

\noindent The full artefact bundle will be released alongside the
de-anonymised camera-ready version of this paper. The inventory
below lists the items the released bundle is committed to contain.

\vspace{4pt}
\textbf{Code and environment.}
\begin{itemize}
\itemsep0pt
\item The evaluation source code (configuration, data loader,
  evaluator, metric suite, bootstrap script, analysis script, and
  the main and BF16-subset entry-points).
\item A \texttt{requirements.txt} with top-level package pins.
  Working versions used for the results in this paper include
  \texttt{torch>=2.1}, \texttt{transformers>=4.40},
  \texttt{bitsandbytes>=0.43}, \texttt{accelerate>=0.27},
  \texttt{datasets>=2.18}, \texttt{scipy>=1.12},
  \texttt{statsmodels>=0.14}, \texttt{pandas>=2.2},
  \texttt{numpy>=1.26}, \texttt{matplotlib>=3.8},
  \texttt{seaborn>=0.13}, and \texttt{huggingface\_hub>=0.20}.
\item Full BitsAndBytes quantisation configuration, the
  \texttt{apply\_chat\_template} flag, and the environment
  lockfile (CUDA runtime, PyTorch build, cuDNN / TF32 / SDPA
  backend configuration).
\item HuggingFace model snapshot pins (commit hashes for each of
  the three model repositories, tokenizer snapshot IDs, and
  per-file SHA-256 of downloaded weights) and the HuggingFace
  dataset revisions for each benchmark.
\end{itemize}

\vspace{4pt}
\textbf{Results, manifests, and figures.}
\begin{itemize}
\itemsep0pt
\item 612 per-cell result records (keyed by configuration, with
  aggregate score, parse rate, and code version), plus the
  matched 54-cell BF16 conservative-core sweep reported in
  Appendix~\ref{appx:bf16}.
\item Metric tables underlying the headline numbers (SDI, CFR,
  ranking, ANOVA, bootstrap) and all figure source files.
\item Per-item and per-seed generation records (raw model text,
  parsed label, gold label, correctness, item ID) for every
  inference call.
\item Stable item manifest (upstream dataset repo, revision, split,
  row index / item ID; 292 items on BBQ, 295 on each of the other
  four benchmarks; subsampling RNG seed).
\item Few-shot exemplar manifest (IDs and order per benchmark and
  per few-shot level, plus exemplar RNG seed).
\item Bootstrap RNG seed and sensitivity-interval type
  (percentile vs.\ BCa / studentised) per reported range.
\end{itemize}

\vspace{4pt}
\textbf{Release-level.}
\begin{itemize}
\itemsep0pt
\item SHA-256 manifest of all released files and a smoke-test
  harness that rebuilds the aggregate numbers for a small subset
  end-to-end from the released code and environment.
\item License declaration (target: CC-BY-4.0 on artefacts
  generated by this project; upstream dataset licenses propagate
  to cached items).
\end{itemize}

\vspace{4pt}
The inventory above is a release commitment, not a claim that
the artefacts are attached at submission time. A
governance-motivated paper about reproducibility should not
pretend to have solved reproducibility at the artefact level when
it has not.

\section{Regex-parser parse rates per (benchmark, template)}
\label{appx:parse}

\noindent Mean and minimum parse rate across all valid free-form
configurations for each (benchmark, template) cell. T4 is
chain-of-thought; parse rates drop materially on T4 for ToxiGen and
XSTest, driving the sensitivity analysis in
§\ref{sec:f1}. CrowS-Pairs is excluded: it admits only the logprob
path, so it has no free-form configurations. (The
$\text{parse\_rate}=1.00$ entries that appeared in earlier
versions of this table for \texttt{crows\_pairs} were trivial
fills from the logprob path, where there is nothing to parse, and
have been removed.)

{\scriptsize
\begin{verbatim}
              T1            T2            T3            T4
benchmark   mean  min    mean  min    mean  min    mean  min    n/cell
bbq         1.00  1.00   1.00  1.00   1.00  1.00   1.00  1.00      36
toxigen     0.98  0.90   1.00  0.99   1.00  0.98   0.88  0.48      36
truthfulqa  1.00  1.00   1.00  1.00   1.00  1.00   1.00  1.00      36
xstest      0.99  0.93   1.00  0.99   1.00  0.98   0.91  0.60      36
\end{verbatim}
}

On the parse-clean subset ($\text{parse\_rate}{\geq}0.95$; 579
of 612 cells), SDI max is unchanged (qwen/truthfulqa
$102.4\%$), $\CFR_{\theta{=}0.5}$ max is unchanged (mistral/truthfulqa
$51.1\%$), $\CFR_{\theta{=}0.7}$ max is unchanged (qwen/bbq
$49.6\%$), $\tauR$ moves by at most $\pm 0.15$ (increasing on
ToxiGen and XSTest), and the $\rho$ shifts are BBQ $3.8{\to}3.8$,
TruthfulQA $0.7{\to}0.7$ (unchanged), ToxiGen $14.9{\to}15.0$
(unchanged), XSTest $2.4{\to}1.7$ (the only slice that moves
materially; still ${>}1$).

\paragraph{Parse-clean rank-flip headlines
(\texttt{rank\_flip\_parse\_clean.py};
\texttt{adversarial\_metrics\_parse\_clean.csv}).} Restricting the
$\rho_{\text{flip}}^{\max}$ / orderings computation to cells with
all-three-models parse rate $\geq 0.95$ leaves the iconic claims
intact: TruthfulQA $\rho_{\text{flip}}^{\max}{=}0.479$ unchanged
on $|C|{=}48$; BBQ $0.438$ on $|C|{=}48$ unchanged; CrowS-Pairs
$0.250$ on $|C|{=}12$ unchanged. ToxiGen drops to
$|C|{=}36$ (12 of 48 cells excluded) and
$\rho_{\text{flip}}^{\max}$ rises slightly to $0.361$. XSTest
drops to $|C|{=}39$ (9 cells excluded) and
$\rho_{\text{flip}}^{\max}$ falls to $0.282$, but the
\textbf{6-of-6 distinct orderings on XSTest is preserved} on this
parse-clean subset. The headline operator-controllable claims
are therefore not contaminated by T4 parse degradation on
ToxiGen / XSTest.

\section{Absolute score ranges per slice}
\label{appx:absranges}

\noindent Scores are in $[0,1]$. ``Range'' is $s_{\max}-s_{\min}$ across
all valid configurations in the (model, benchmark) slice. Columns: $n$
is the number of valid configs in the slice, then $s_{\min}$,
$s_{\max}$, $\bar s$, and absolute range.

{\footnotesize
\begin{verbatim}
                          n     min     max    mean   range
mistral / bbq            48   0.399   0.764   0.661   0.365
mistral / crows_pairs    12   0.247   0.329   0.288   0.081
mistral / toxigen        48   0.505   0.881   0.798   0.376
mistral / truthfulqa     48   0.373   0.695   0.519   0.322
mistral / xstest         48   0.505   0.933   0.851   0.428
qwen    / bbq            48   0.331   0.860   0.694   0.529
qwen    / crows_pairs    12   0.180   0.254   0.215   0.075
qwen    / toxigen        48   0.505   0.875   0.811   0.369
qwen    / truthfulqa     48   0.140   0.693   0.540   0.553
qwen    / xstest         48   0.505   0.926   0.803   0.421
yi      / bbq            48   0.342   0.840   0.681   0.499
yi      / crows_pairs    12   0.288   0.349   0.325   0.061
yi      / toxigen        48   0.505   0.854   0.757   0.349
yi      / truthfulqa     48   0.369   0.732   0.592   0.363
yi      / xstest         48   0.505   0.898   0.761   0.393
\end{verbatim}
}

\section{ANOVA specification}
\label{appx:anova}

The variance decomposition in §\ref{sec:f1} is computed, for each
benchmark, on the free-form scoring subset of that benchmark (so
\texttt{scoring} is a constant in the slice). Quantization is also a
constant (\texttt{nf4}). The fitted ordinary-least-squares model is

\begin{quote}\footnotesize
\texttt{score \textasciitilde{} C(model) + C(template) +
C(decoding) + C(few\_shot) \\ + C(model):C(template) +
C(model):C(decoding) + C(model):C(few\_shot) \\ +
C(template):C(decoding) + C(template):C(few\_shot)}
\end{quote}

Two variance-share estimators are computed from the same Type-II
ANOVA table.

\begin{figure}[H]
\centering
\includegraphics[width=\columnwidth]{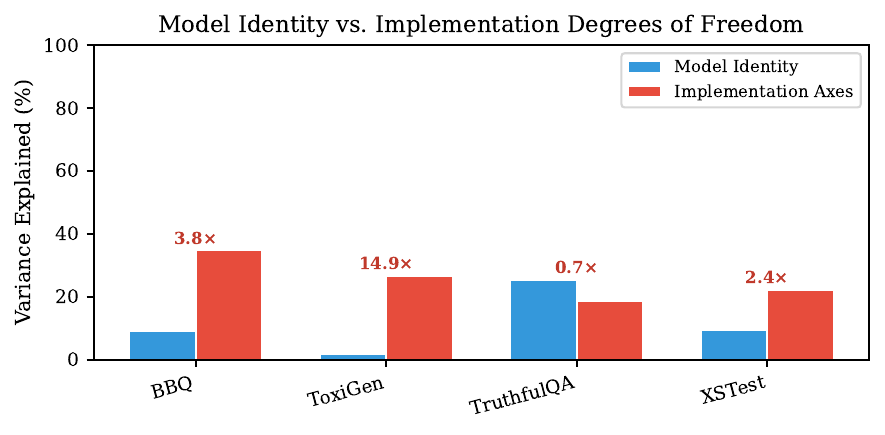}
\caption{Variance decomposition; $\rho{>}1$ on BBQ/ToxiGen/XSTest,
$\rho{<}1$ on TruthfulQA.}
\label{appx:fig-mvi}
\end{figure}

The headline decomposition in Table~\ref{tab:mvi} and Fig.~\ref{appx:fig-mvi}
(the ratio $\rho$) uses the raw-$\eta^2$ share
$\eta^2_X = SS_X / SS_\text{total}$. The \emph{model} share is
$\eta^2_{\text{model}}$. The \emph{implementation} share is
$\eta^2_{\text{template}} + \eta^2_{\text{decoding}} +
\eta^2_{\text{few-shot}}$. The \emph{interaction} share is the sum
of $\eta^2$ over all included two-way interaction terms
(including model$\times$*). The implementation-to-model ratio
$\rho = \eta^2_{\text{impl}} / \eta^2_{\text{model}}$. Under this
convention the four columns (model / impl. / inter. / residual) sum
to $100\%$ by construction.

The per-axis breakdown in Fig.~\ref{fig:anova} uses the error-corrected
Hays' $\omega^2$~\citep{hays1994statistics} instead:
\[
  \omega^2_X \;=\; \max\!\Big(\tfrac{SS_X - df_X \cdot \mathrm{MS}_W}{SS_\text{total} + \mathrm{MS}_W},\;0\Big),
\]
which penalises terms whose sum of squares is consistent with
residual noise. This is a better summary when the question is
``which axis alone explains measurable score variance'' rather than
``how does variance partition overall''.

CrowS-Pairs is excluded from this analysis because its free-form
slice is empty. The analysis is a deliberate first-pass for reasons
given in §\ref{sec:metrics}.

\section{Within-grid configuration-bootstrap sensitivity intervals}
\label{appx:boot}

\noindent The table below is extracted from
\texttt{results/metric\_bootstrap\_summary.md}. Each row is 1{,}000
bootstrap resamples over configurations.

\vspace{4pt}
\noindent\textbf{SDI} (per model/benchmark, \%):
{\footnotesize
\begin{verbatim}
group                 point   mean    lo     hi
mistral/bbq           55.2    53.2    25.6   57.1
mistral/crows_pairs   28.2    27.5    15.8   29.8
mistral/toxigen       47.1    41.7    29.9   48.6
mistral/truthfulqa    62.0    60.3    49.9   64.0
mistral/xstest        50.3    41.4    23.2   51.7
qwen/bbq              76.3    75.4    68.4   82.7
qwen/crows_pairs      34.7    33.9    20.2   37.0
qwen/toxigen          45.5    40.0    29.4   46.7
qwen/truthfulqa      102.4    99.2    56.5  109.1
qwen/xstest           52.5    50.3    45.8   54.3
yi/bbq                73.2    71.7    63.4   78.6
yi/crows_pairs        18.8    18.2     6.0   19.5
yi/toxigen            46.2    46.0    43.9   47.9
yi/truthfulqa         61.3    59.3    48.3   63.7
yi/xstest             51.7    51.2    48.5   53.5
\end{verbatim}
}

\vspace{4pt}
\noindent\textbf{$\CFR_{\theta=0.5}$} (per model/benchmark, \%).
CrowS-Pairs rows are marked N/A because a fixed compliance
threshold is not semantically meaningful for its
stereotype-preference score (no conventional pass/fail cutoff),
and within this grid it admits only the logprob scoring path; the
blanks should not be read as evidence of stability:
{\footnotesize
\begin{verbatim}
group                 point   mean    lo     hi
mistral/bbq           12.0    11.4     0.0   25.4
mistral/crows_pairs    N/A     N/A    N/A    N/A
mistral/toxigen        0.0     0.0     0.0    0.0
mistral/truthfulqa    51.1    50.0    45.4   51.1
mistral/xstest         0.0     0.0     0.0    0.0
qwen/bbq              25.4    24.4    12.0   38.3
qwen/crows_pairs       N/A     N/A    N/A    N/A
qwen/toxigen           0.0     0.0     0.0    0.0
qwen/truthfulqa       38.3    37.1    22.3   47.9
qwen/xstest            0.0     0.0     0.0    0.0
yi/bbq                43.9    43.1    31.1   50.7
yi/crows_pairs         N/A     N/A    N/A    N/A
yi/toxigen             0.0     0.0     0.0    0.0
yi/truthfulqa         28.4    27.7    12.0   40.3
yi/xstest              0.0     0.0     0.0    0.0
\end{verbatim}
}

\vspace{4pt}
\noindent\textbf{$\CFR_{\theta=0.7}$} (per model/benchmark, \%). CrowS-Pairs N/A as above:
{\footnotesize
\begin{verbatim}
group                 point   mean    lo     hi
mistral/bbq           42.2    41.2    28.4   49.6
mistral/crows_pairs    N/A     N/A    N/A    N/A
mistral/toxigen       25.4    24.8    12.0   38.3
mistral/truthfulqa     0.0     0.0     0.0    0.0
mistral/xstest         8.2     7.9     0.0   19.1
qwen/bbq              49.6    48.6    40.3   51.1
qwen/crows_pairs       N/A     N/A    N/A    N/A
qwen/toxigen          15.6    14.9     4.2   28.4
qwen/truthfulqa        0.0     0.0     0.0    0.0
qwen/xstest           19.1    19.1     4.2   33.7
yi/bbq                43.9    43.1    31.1   50.7
yi/crows_pairs         N/A     N/A    N/A    N/A
yi/toxigen            22.3    22.2     8.2   36.1
yi/truthfulqa         22.3    22.0     8.2   36.1
yi/xstest             22.3    22.0     8.2   36.1
\end{verbatim}
}

\vspace{4pt}
\noindent\textbf{$\tauR$} (benchmark: point/mean/lo/hi):
{\footnotesize
\begin{verbatim}
bbq          0.141 / 0.141 / -0.059 / 0.487
crows_pairs  0.632 / 0.646 /  0.000 / 1.000
toxigen      0.128 / 0.131 / -0.067 / 0.418
truthfulqa   0.053 / 0.053 / -0.079 / 0.297
xstest       0.158 / 0.158 / -0.067 / 0.492
\end{verbatim}
}

\vspace{4pt}
\noindent\textbf{Variance attribution} (\%, point/mean/lo/hi):
{\footnotesize
\begin{verbatim}
benchmark      model                   impl                     rho point
bbq             9.1 / 10.6 / 3.1/23.7  34.7 / 36.9 /19.1/51.4    3.8
toxigen         1.8 /  4.0 / 0.1/13.2  26.6 / 28.6 /17.7/39.5   14.9
truthfulqa     25.4 / 26.5 /14.8/42.7  18.7 / 21.9 /11.3/33.3    0.7
xstest          9.3 / 11.6 / 2.7/24.3  22.2 / 24.6 /13.1/37.6    2.4
\end{verbatim}
}

\section{Type-II vs Type-III variance attribution}
\label{appx:typeIII}

As a robustness check on the headline variance-ratio claim, we refit
the same four-way ANOVA on the free-form slice under both Type-II and
Type-III sums of squares (the paper uses Type-II). Implementation share
is the sum of the template, decoding and few-shot $\eta^2$ shares;
$\rho=\text{impl share}/\text{model share}$.

{\footnotesize
\begin{verbatim}
benchmark      T-II rho  model T-II  impl T-II  |  T-III rho  model T-III  impl T-III
bbq                3.83      9.1%       34.7%   |     3.28        9.1%        29.8%
toxigen           14.91      1.8%       26.6%   |     4.30        1.8%         7.7%
truthfulqa         0.74     25.4%       18.7%   |     0.79       25.4%        20.0%
xstest             2.39      9.3%       22.2%   |     1.33        9.3%        12.4%
\end{verbatim}
}

Under Type-III the qualitative ordering survives: implementation
dominates on BBQ and ToxiGen ($\rho{\gg}1$), model dominates on
TruthfulQA ($\rho{<}1$), and XSTest stays implementation-close
($\rho=1.33$, model 9.3\% / impl 12.4\%) rather than flipping
sides --- the magnitude shrinks substantially relative to Type-II
($\rho=2.39$) but does not cross unity. The benchmark-level
sensitivity on XSTest should be read as a caution against
over-interpreting any single benchmark's $\rho$ magnitude rather
than as a sign-change.

\section{Multiplicity-controlled headline maxima}
\label{appx:multiplicity}

We report 15 (model, benchmark) $\SDI$s, up to 60 $\CFR$s, 5
$\tauR$s, and 4 variance decompositions; ``up to $X$'' statements
in the body are maxima over $N$ slices. To bound the multiplicity
inflation, within each non-parametric configuration bootstrap
\citep{efron1993bootstrap} we recompute the metric on every slice
and take the max; the 2.5 / 97.5 percentiles of the resulting max
distribution form the max-statistic resampling interval over the
observed grid. SDI max: $102.4\%$,
interval $[76.1,\,109.1]\%$; $\CFR_{0.5}$ max: $51.1\%$, interval
$[46.7,\,51.1]\%$; $\CFR_{0.7}$ max: $49.6\%$, interval
$[43.9,\,51.1]\%$; $\rho$ max: $14.9$, interval $[3.2,\,249.2]$.
Every covered headline max (SDI, $\CFR$, $\rho$) remains
substantively large under max-statistic bootstrap resampling, so
the qualitative narrative for those metrics is not driven by
multiplicity. The $\rho_{\text{flip}}^{\max}$ headline is excluded
from this multiplicity correction by construction (exact counts on
the observed envelope, no inferential maximisation).

\section{Supporting figures}
\label{appx:extra-figs}

\begin{figure}[t]
\centering
\includegraphics[width=0.5\textwidth]{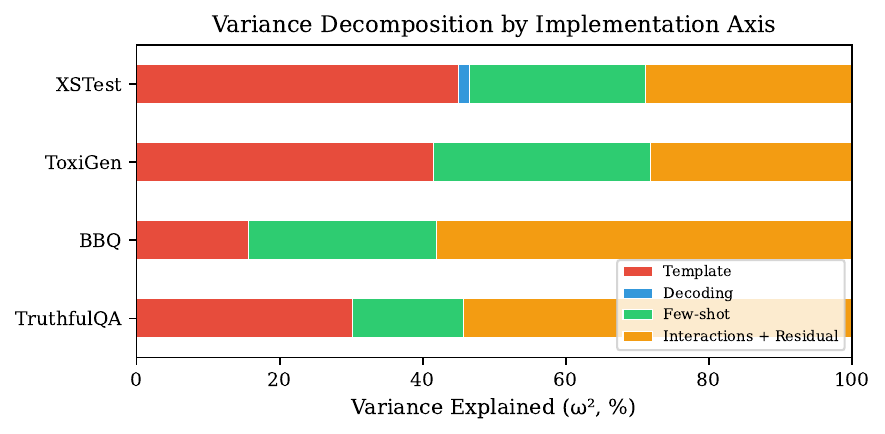}
\caption{No single implementation axis dominates on any benchmark,
so one-at-a-time robustness sweeps under-report the joint envelope.
Per-axis $\omega^2$ shares within the implementation column of
Table~\ref{tab:mvi}.}
\label{fig:anova}
\end{figure}

\begin{figure}[t]
\centering
\includegraphics[width=0.5\textwidth]{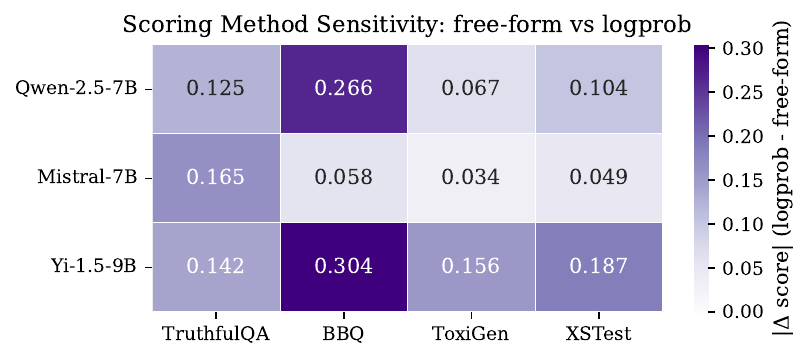}
\caption{Switching only the scoring pipeline (free-form regex parse
$\leftrightarrow$ logprob argmax), with everything else held fixed.
Cells show the \emph{mean} absolute score gap across matched greedy
configurations per (model, benchmark); the \emph{max} per cell is
larger (e.g.\ 0.468 on Qwen-2.5-7B / TruthfulQA, source
\texttt{metric\_scoring\_method\_effect.csv}). CrowS-Pairs excluded
because it supports only the logprob path.}
\label{fig:scoring}
\end{figure}

\section{Per-benchmark summary}
\label{appx:benchmarks}

\begin{description}
\itemsep0pt
\item[TruthfulQA] Highest SDI (up to $102.4\%$ on Qwen-2.5-7B); the
  only benchmark where model identity dominates ($\rho=0.7$);
  $\tauR{\approx}0$, consistent with chance; scoring-method swings
  up to $0.468$. Leaderboard order between these three models under
  this grid carries essentially no information beyond the harness
  choice.
\item[BBQ] Implementation-dominated ($\rho=3.8\times$);
  $\CFR_{0.5}$ up to $43.9\%$ on Yi-1.5-9B and $\CFR_{0.7}$ up to
  $49.6\%$ on Qwen-2.5-7B (the overall $\CFR_{0.7}$ maximum);
  scoring-method sensitive (Qwen: $|\Delta|=0.266$).
\item[ToxiGen] Highest $\rho$ overall ($14.9\times$, within-grid sensitivity range $[3.2,\,249.2]$ --- the wide range is essential context; see \S\ref{sec:f1}) because the
  three models land in a narrow high-accuracy band, so the
  model-variance share is only $1.8\%$ and any implementation
  movement dominates; total-order mismatch rate $78.3\%$, near but
  not above the $83.3\%$ chance baseline (the $M{=}3$ uniform null
  for total-order mismatch is $5/6$); interaction variance dominant.
\item[XSTest] Implementation-dominated under Type-II ($\rho=2.4$)
  but the most method-convention-sensitive benchmark: under
  Type-III SS $\rho$ drops to $1.3$ (Appendix~\ref{appx:typeIII})
  and on the parse-clean subset to $1.7$ (§\ref{sec:f1}),
  so XSTest is best read as boundary-close rather than cleanly
  impl-dominant. Non-trivial $\CFR_{0.7}$ on Yi/XSTest ($22.3\%$).
\item[CrowS-Pairs] Low SDI, and $\CFR$ not defined (N/A) by
  construction: the pair-format admits only one scoring path and
  one decoding, so the fixed-threshold flip rate is algebraically
  degenerate rather than informative.
\end{description}

\section{Per-cell $\CFR$ tables and median-panel ceiling}
\label{appx:cfr-tables}

\paragraph{All four fixed thresholds plus median.} Full $\CFR$
tables for $\theta\in\{0.5,0.6,0.7,0.8\}$ and at the per-cell
median, in percentages, v3.0 point estimates. CrowS-Pairs rows are
N/A because the pair format admits only the logprob scoring path,
collapsing the grid to one scoring instance; a single-threshold
$\CFR$ is not semantically meaningful for the
stereotype-preference score (no conventional pass/fail cutoff).

{\footnotesize
\begin{verbatim}
CFR at theta=0.5     BBQ   CrowS  ToxiGen  TruthQA  XSTest
Mistral             12.0    N/A     0.0     51.1     0.0
Qwen                25.4    N/A     0.0     38.3     0.0
Yi                  43.9    N/A     0.0     28.4     0.0

CFR at theta=0.6     BBQ   CrowS  ToxiGen  TruthQA  XSTest
Mistral             15.6    N/A     4.2     33.7     4.2
Qwen                47.9    N/A     4.2     50.7    15.6
Yi                  43.9    N/A    22.3     51.1    22.3

CFR at theta=0.7     BBQ   CrowS  ToxiGen  TruthQA  XSTest
Mistral             42.2    N/A    25.4      0.0     8.2
Qwen                49.6    N/A    15.6      0.0    19.1
Yi                  43.9    N/A    22.3     22.3    22.3

CFR at theta=0.8     BBQ   CrowS  ToxiGen  TruthQA  XSTest
Mistral              0.0    N/A    47.9      0.0    36.1
Qwen                50.3    N/A    28.4      0.0    46.7
Yi                  38.3    N/A    50.3      0.0    48.8

CFR at per-cell      BBQ   CrowS  ToxiGen  TruthQA  XSTest
median (=ceiling)
Mistral             51.1    N/A    51.1     51.1    51.1
Qwen                51.1    N/A    51.1     51.1    51.0
Yi                  51.1    N/A    51.1     51.1    51.1
\end{verbatim}
}

\paragraph{Median-panel algebraic ceiling.} The median panel sits
at $\approx\!51\%$ uniformly because, by the identity
$\CFR_\theta=\tfrac{2n}{n-1}\,p_\theta(1-p_\theta)$, choosing
$\theta$ as the per-cell median forces the pass-fraction
$p_\theta$ to ${\approx}1/2$ on every non-degenerate slice, which
pins $\CFR$ to the $p(1-p)$ maximum
$\tfrac{2n}{n-1}\cdot\tfrac{1}{4}$. For $n{=}48$ (the free-form
non-pair cells) the ceiling is $51.06\%$; for $n{=}12$ (the logprob
slice) it would be $54.55\%$. The median panel therefore reports
an algebraic ceiling, not a substantive finding, which is why the
body uses only $\theta\in\{0.5,0.7\}$ as illustrative procurement-
style thresholds.

\section{Eight bugs fixed in the v3.0 rerun}
\label{appx:code-audit}

The eight correctness issues surfaced by the adversarial code review, in
decreasing order of empirical impact. Quantitative confirmation of the
impact ordering is in Appendix~\ref{appx:bugfix-impact}.

\textbf{(B2, high impact).} Tokenizer truncation defaulted to the
right side while \texttt{max\_length}${=}2048$ was active. On
few-shot prompts whose chat-templated length exceeded the cap, the
v2.2 harness silently right-truncated the test question and the
\texttt{Assistant:} cue out of the model's view. The v3.0 fix
sets \texttt{tokenizer.truncation\_side}=\texttt{left}, which
preserves the question at the expense of the first exemplars.
Empirically, v3.0 per-configuration wall-clock rose from
${\sim}180\,$s to ${\sim}430\,$s average because the model now
actually processes the intended prompt length; the 5-shot
configurations were the most affected.
\textbf{(B4, modest impact).} The free-form binary answer extractor
scanned only the first 64--96 characters for yes/no and safe/unsafe
verdicts; T4 was protected via a whole-response fallback, so the
material effect is on T1/T2 parse rates (understated by ${\leq}3$\%
pre-fix). The v3.0 extractor scans the full response with a
last-explicit-answer heuristic.
\textbf{(B7, ratio-preserving).} Hays' $\omega^2$ used a
non-standard denominator in v2.2; v3.0 uses Hays' $\omega^2$ with
the canonical $\mathrm{SS}_\text{total}+\mathrm{MS}_W$
normalisation defined in App.~\ref{appx:anova} (where
$\mathrm{MS}_W$ is the within-cell mean square; we standardise
``$\mathrm{MS}_\text{res}$'' to ``$\mathrm{MS}_W$'' across the
paper to match the App.~\ref{appx:anova} formula).
The per-axis bar heights in Fig.~\ref{fig:anova} move slightly; the
ratio $\rho$ in Table~\ref{tab:mvi} is an $\eta^2$ share
(Appendix~\ref{appx:anova}) and is unaffected.
\textbf{(B8, benchmark-scoped).} ToxiGen labels fell back to an
AI-classifier score when the human-annotator mean was missing;
v3.0 restricts labels to \texttt{toxicity\_human} and drops rows
with missing annotations.
\textbf{(B1, direction of CrowS only).} A type mismatch on the
CrowS-Pairs \texttt{stereo\_antistereo} field inverted the per-item
stereotype direction on $84\%$ of rows; because the reported CrowS
$\SDI$, $\CFR$, $\tauR$ and $\rho$ are derived from a
direction-agnostic bias-magnitude score, those metrics are
unchanged and only the auxiliary \texttt{stereo\_pref\_mean} moved.
\textbf{(B3, argmax-robust).} Chat-templated prompts were
re-tokenized with default \texttt{add\_special\_tokens=True}, which
could double-insert BOS on Mistral/Yi; for argmax-based logprob
scoring this adds a constant penalty across candidates and is
robust, though absolute log-likelihoods were off by that constant.
\textbf{(B5, rendering-equivalent).} The CrowS-Pairs pair template
carried a never-substituted \texttt{\{sent\}} placeholder; the
rendered output was byte-identical to the fixed version.
\textbf{(B6, untriggered).} Empty-continuation rows would have
fallen back to a surrogate UNK/EOS token; no real (model,
benchmark) cell in our grid triggered this path.
The v3.0 headline maxima reported in §\ref{sec:findings} are
$\SDI=102.4\%$ on Qwen/TruthfulQA, $\CFR_{0.5}=51.1\%$ on
Mistral/TruthfulQA, $\CFR_{0.7}=49.6\%$ on Qwen/BBQ, $\tauR=0.05$
on TruthfulQA, and $\rho\in\{3.8, 14.9, 0.7, 2.4\}$ on (BBQ,
ToxiGen, TruthfulQA, XSTest).

\section{v2.2$\to$v3.0 bugfix impact}
\label{appx:bugfix-impact}

Because the v2.2 code-version result JSONs are preserved in git, we
can measure the bugfix impact as a paired delta without an additional
GPU run. For each of the 612 result cells we pair the v2.2 score
(from the v2.2 commit, hash redacted for blind review) against the v3.0 score and report the
delta $\Delta = s_\text{v3.0}-s_\text{v2.2}$ aggregated along the
axes where B2 and B4 were expected to bite.

{\footnotesize
\begin{verbatim}
axis    level       N     mean Delta     sd      max |Delta|
few_shot fs0        204   -0.012        0.071     0.403
few_shot fs3        204   +0.015        0.094     0.489
few_shot fs5        204   +0.016        0.089     0.506
template T1         153   -0.003        0.041     0.148
template T2         153   -0.023        0.065     0.403
template T3         153   +0.012        0.075     0.347
template T4         153   +0.039        0.127     0.506
model    mistral    204   -0.010        0.033     0.207
model    qwen       204   -0.006        0.029     0.160
model    yi         204   +0.035        0.138     0.506
\end{verbatim}
}

Individual-cell maxima reach 0.51 absolute score points. The fs5 /
T4 corner, where B2 (right-truncation) and B4 (regex head-only) both
had the most room to bite, accounts for most of the tail. Yi-1.5 is
the most bugfix-sensitive model on average; Qwen and Mistral are
essentially inside the v2.2 noise band on average but have tail cells
that moved substantially. This is the direct truncation-side /
parser-scope ablation called for in the priority follow-ups (\S\ref{sec:threats}, item viii).

\section{BF16 conservative-core sweep (NF4 vs BF16, paired cells)}
\label{appx:bf16}

\paragraph{Sweep design.} On the RTX PRO 6000 Blackwell (96\,GB) we
re-ran the §\ref{sec:f-adversarial} conservative-core
sub-envelope (T1+T3 $\times$ greedy $\times$ 0-shot $\times$
\{free-form, logprob\}) under BF16 precision for all three 7--9B
models and all five benchmarks --- the same envelope we used for
NF4 in the main grid. This produces 54 paired NF4/BF16 cells, one
per (model, benchmark, template, scoring) combination. The 16\,GB
constraint that previously forced dropping Yi-1.5-9B from the BF16
subset no longer applies on this hardware. Parse rate on the
free-form path is $\geq 96\%$ on every cell.

\paragraph{Per-cell delta summary (54 paired cells).}
The mean $|s_{\text{BF16}}-s_{\text{NF4}}|$ is $2.58$ pp, median
$1.70$ pp, max $20.68$ pp. $9.3\%$ of cells (5 of 54) exceed a
$5$-pp gap; $3.7\%$ (2 cells) exceed $10$ pp. By model: Yi-1.5-9B
is the outlier (mean $|\Delta|=3.8$ pp, max $20.68$ pp on
Yi/XSTest/T1/logprob); Qwen-2.5-7B and Mistral-7B are tighter
(mean $|\Delta|\approx 2$ pp, max $\approx 7.2$ pp).
Five largest paired deltas (NF4 vs BF16, percentage points):

{\footnotesize
\begin{verbatim}
model     bench      tpl  scoring     NF4    BF16   Delta
yi        xstest     T1   logprob    70.5   49.8   -20.7
yi        toxigen    T1   logprob    71.9   53.9   -18.0
qwen      truthfulqa T3   free_form  63.6   70.8   +7.2
mistral   toxigen    T3   free_form  62.7   69.8   +7.1
qwen      toxigen    T1   free_form  61.7   67.1   +5.4
\end{verbatim}
}

\paragraph{What this maps.} Under matched conservative-core configs,
\emph{most} cells move by ${<}3$ pp under the NF4$\to$BF16 swap;
the within-envelope conservative-core score-range spread therefore
\emph{persists} at BF16 on most (model, benchmark) cells (and on
two cells, Yi/ToxiGen and Yi/XSTest, the BF16 within-envelope
range is in fact \emph{larger} than the NF4 range). We present
this as paired score-range evidence on the conservative core, and
provide the matching conservative-core paired-ordering result
below (94.4\% match). We do not extend a paired ordering claim
to the full $|C|{=}48$ adversarial envelope: a 1--3\,pp paired
score-range delta is a comparability statement on score
\emph{magnitudes}, not on pairwise model orderings outside the
conservative-core slice, and the
54-cell sweep does not include matched A/B pair-flip counts.
The paired NF4/BF16 ordering-stability table on the
conservative-core sub-envelope follows below (94.4\%, $17/18$
cells); we do not extend that ordering claim to the full
$|C|{=}48$ adversarial envelope, where pairwise verdicts could
respond differently to an NF4$\to$BF16 swap on configurations we
have not paired.
\emph{But} the tail is heavy --- the single largest delta in the
54-cell sweep is 20.7 pp on Yi/XSTest/T1/logprob, with five total
cells exceeding a 5\,pp NF4$\to$BF16 swing (top-five table above). The
within-envelope conservative-core score range itself is comparable
in magnitude under both precisions on most (model, benchmark)
cells, with the notable exception of Yi on ToxiGen and XSTest
where BF16 produces a \emph{larger} within-envelope range than NF4
(Yi/ToxiGen $0.102{\to}0.319$; Yi/XSTest $0.139{\to}0.342$).
Per-(model, benchmark) abs ranges:

{\footnotesize
\begin{verbatim}
model     benchmark    NF4 range  BF16 range
mistral   bbq          0.134      0.110
mistral   crows_pairs  0.047      0.061
mistral   toxigen      0.223      0.153
mistral   truthfulqa   0.207      0.214
mistral   xstest       0.105      0.102
qwen      bbq          0.445      0.473
qwen      crows_pairs  0.014      0.007
qwen      toxigen      0.217      0.183
qwen      truthfulqa   0.494      0.597
qwen      xstest       0.346      0.356
yi        bbq          0.411      0.425
yi        crows_pairs  0.020      0.007
yi        toxigen      0.102      0.319
yi        truthfulqa   0.217      0.217
yi        xstest       0.139      0.342
\end{verbatim}
}

\begin{figure}[H]
\centering
\includegraphics[width=\columnwidth]{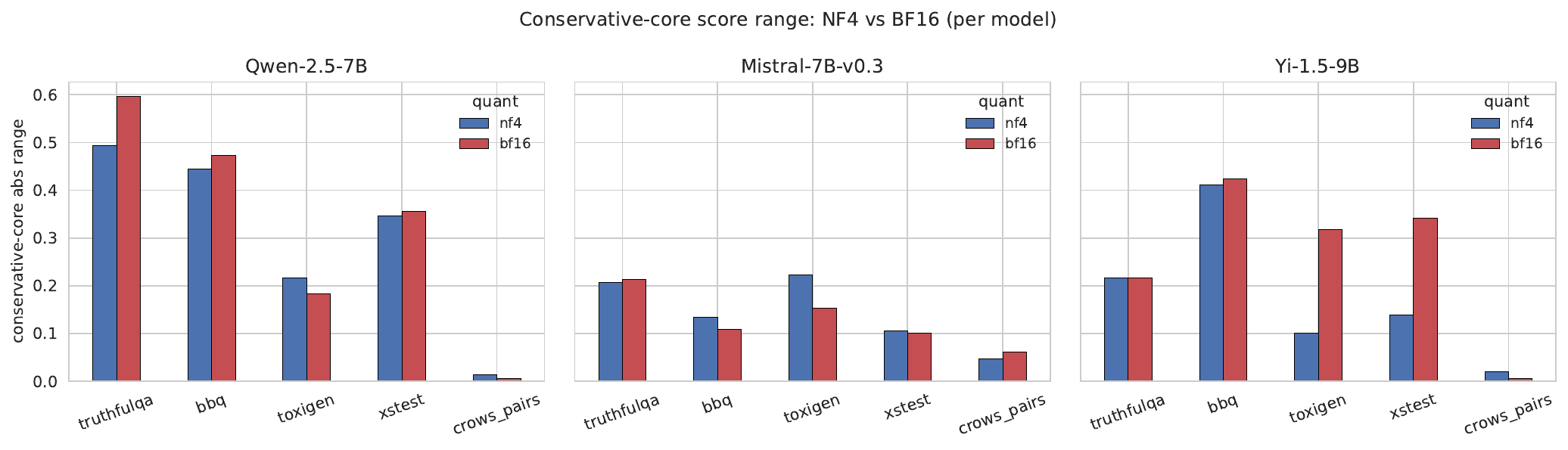}
\caption{Conservative-core score range (max-min) per
(model, benchmark) under NF4 (blue) vs BF16 (red). The
within-envelope spread persists under both precisions for Qwen and
Mistral; Yi-1.5-9B picks up extra spread on ToxiGen / XSTest at
BF16, indicating a precision-by-model interaction worth tracking.}
\label{fig:bf16-cc}
\end{figure}

\paragraph{Reading.} Precision is no longer ``incompletely swept''
for the conservative core: every (model, benchmark, template,
scoring) cell has a matched NF4/BF16 partner. The result is that
precision deltas are median-small but tail-risky under matched
conservative-core configs (median 1.7 pp, tail max 20.7 pp), with
the heavy tail concentrated on Yi-1.5-9B in particular. Looking
at the per-cell breakdown, the two largest deltas
(Yi/XSTest/T1/logprob $-20.7$ pp and Yi/ToxiGen/T1/logprob
$-18.0$ pp) are on the \emph{logprob} scoring path, not free-form;
the next three largest deltas (Qwen/TruthfulQA/T3, Mistral/ToxiGen/T3,
Qwen/ToxiGen/T1) are on the free-form path with magnitudes
$5{-}7.2$ pp. The precision axis is therefore \emph{not}
median-attributable to a single scoring path; the heavy tail on
Yi is concentrated on T1/logprob cells. A full BF16 sweep across
the other implementation axes (decoding, few-shot, T2/T4) on
$\geq$24\,GB hardware is the natural extension; this work
establishes the conservative-core baseline.

\paragraph{Paired NF4/BF16 ordering stability (18 cells).} Computing
the induced 3-model ordering at each (benchmark, template, scoring)
cell on both precisions
(\texttt{bf16\_ordering\_stability.py}; output
\texttt{bf16\_ordering\_stability.csv}) gives \textbf{17 of 18}
cells (94.4\%) where the NF4 ordering matches the BF16 ordering
exactly with zero pair flips. The single divergent cell is
ToxiGen / T3 / logprob, where the (Qwen, Yi) pair flips between
NF4 (qwen $>$ yi $>$ mistral) and BF16
(yi $>$ qwen $>$ mistral); Mistral remains last. This single
flip occurs on the same benchmark and scoring path as the
heaviest-tail Yi/ToxiGen/T1/logprob $-18.0$ pp cell, but at
T3/logprob, where the (Qwen, Yi) margin is small. On the conservative-core sub-envelope, then,
NF4$\to$BF16 ordering stability is $94.4\%$ ($17/18$ cells), with
\emph{no} cells exhibiting a full ($\geq 2$-pair) reversal. The
remaining gap to a paired ordering-stability claim on the full
$|C|{=}48$ adversarial envelope is the missing T2/T4 / 3-shot /
5-shot / diverse-decoding cells under BF16; we report the
conservative-core paired-ordering result here and leave the
broader sweep to future work.

\section{Per-cell SDI table and supporting variance figure}
\label{appx:sdi-extra}

\begin{table}[H]
\centering\small
\caption{Per-(model, benchmark) SDI (\%), point estimates;
configuration-bootstrap sensitivity intervals in App.~\ref{appx:boot}.}
\label{tab:sdi}
\begin{tabular}{lrrrrr}
\toprule
 & BBQ & CrowS & ToxiGen & TruthQA & XSTest \\
\midrule
Mistral & 55.2 & 28.2 & 47.1 & 62.0 & 50.3 \\
Qwen    & 76.3 & 34.7 & 45.5 & 102.4 & 52.5 \\
Yi      & 73.2 & 18.8 & 46.2 & 61.3 & 51.7 \\
\bottomrule
\end{tabular}
\end{table}

\begin{figure}[H]
\centering
\includegraphics[width=\columnwidth]{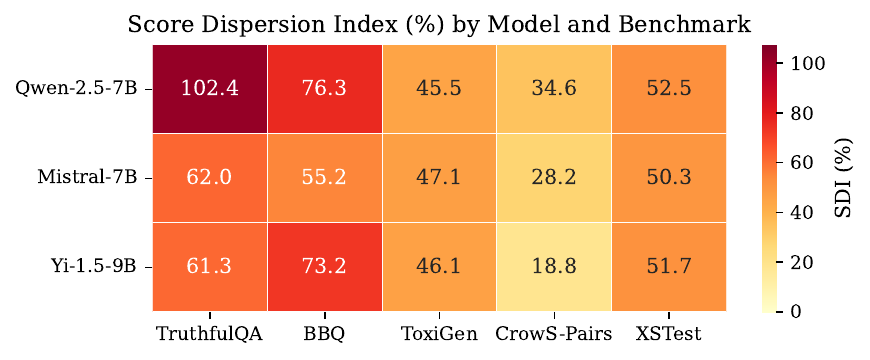}
\caption{SDI heatmap, $(\max\!-\!\min)/\bar s$ per (model, benchmark).}
\label{fig:sdi}
\end{figure}

\section{Cross-family scale probe (Qwen-2.5 7B$\to$32B + Yi-1.5 9B$\to$34B)}
\label{appx:scale-probe}

We re-ran the conservative-core sub-envelope (T1 + T3, greedy,
0-shot, free-form + logprob, NF4) at the larger sibling of two
families already in the §\ref{sec:f-adversarial} grid:
Qwen-2.5-32B-Instruct (paired with Qwen-2.5-7B-Instruct) and
Yi-1.5-34B-Chat (paired with Yi-1.5-9B-Chat). All large-model runs
use the same fixed item subsets as the main grid (292 BBQ items, 295 on each other benchmark, seed 42). This
is a \emph{two-family, four-point} probe; it is not a scale axis
and we still do not claim a generic scaling trend across the
open-weight frontier. What it does do is rule out the
single-family confound: the pattern below is shared across Qwen-2.5
and Yi-1.5, not a Qwen-specific artefact. We deliberately keep
this posture narrower than work that \emph{posits} a generic
scaling law (e.g.,~\citet{shi2024scaling} for time-series
forecasting): the two-family four-point probe rules out a confound
but does not license a general scaling-law claim.

\begin{table}[H]
\centering\footnotesize
\caption{Conservative-core SDI (\%) and absolute score range
($s_{\max}{-}s_{\min}$) for the small/large pair of two model
families, same anchor (greedy, 0-shot, NF4) over the T1+T3
$\times$ \{free-form, logprob\} sub-envelope (4 configs per
free-form benchmark; 2 for CrowS-Pairs which is logprob-only).}
\label{tab:scale}
\setlength{\tabcolsep}{3pt}
\begin{tabular}{@{}lrrrrrrrr@{}}
\toprule
& \multicolumn{4}{c}{Qwen-2.5} & \multicolumn{4}{c}{Yi-1.5} \\
\cmidrule(lr){2-5} \cmidrule(lr){6-9}
& \multicolumn{2}{c}{SDI (\%)} & \multicolumn{2}{c}{range} & \multicolumn{2}{c}{SDI (\%)} & \multicolumn{2}{c}{range} \\
benchmark   & 7B    & 32B   & 7B    & 32B   & 9B    & 34B  & 9B    & 34B  \\
\midrule
TruthfulQA  &  99.3 & 113.8 & 0.494 & 0.631 &  40.3 & 44.6 & 0.217 & 0.275 \\
BBQ         &  78.0 &  81.2 & 0.445 & 0.462 &  70.1 & 67.0 & 0.411 & 0.442 \\
ToxiGen     &  28.0 &   4.3 & 0.217 & 0.037 &  12.9 &  4.9 & 0.102 & 0.042 \\
XSTest      &  47.2 &   7.8 & 0.346 & 0.068 &  17.7 &  2.4 & 0.139 & 0.020 \\
CrowS-Pairs &   6.3 &   0.0 & 0.014 & 0.000 &   6.0 &  0.0 & 0.020 & 0.000 \\
\bottomrule
\end{tabular}
\end{table}

\paragraph{What the probe shows.} The same three-vs-two split
appears in both families: \emph{ToxiGen, XSTest, CrowS-Pairs}
narrow to near-zero conservative-core range at the larger model in
each family (Qwen-2.5: $0.217{\to}0.037$, $0.346{\to}0.068$,
$0.014{\to}0.000$; Yi-1.5: $0.102{\to}0.042$, $0.139{\to}0.020$,
$0.020{\to}0.000$), while \emph{TruthfulQA and BBQ} keep --- or
slightly grow --- their within-envelope range at the larger model
(Qwen-2.5 TruthfulQA $0.494{\to}0.631$, BBQ $0.445{\to}0.462$;
Yi-1.5 TruthfulQA $0.217{\to}0.275$, BBQ $0.411{\to}0.442$). The
absolute SDI levels differ between families (Yi-1.5 is overall
lower-variance than Qwen-2.5 on this sub-envelope), but the
\emph{which-benchmarks-collapse-vs-persist} partition is consistent.

\paragraph{What it still does not show.} Two families with two
scale points each is sufficient to rule out the single-family
confound but not a scaling trend across the open-weight frontier;
we do not claim transfer to Mistral, Llama, closed-weight frontier
models, or the full §\ref{sec:f-adversarial} envelope. A
Llama-3.1-70B run was attempted but excluded by access gating; a
third family or a within-family multi-point sweep is the natural
next step.

\begin{figure}[H]
\centering
\includegraphics[width=\columnwidth]{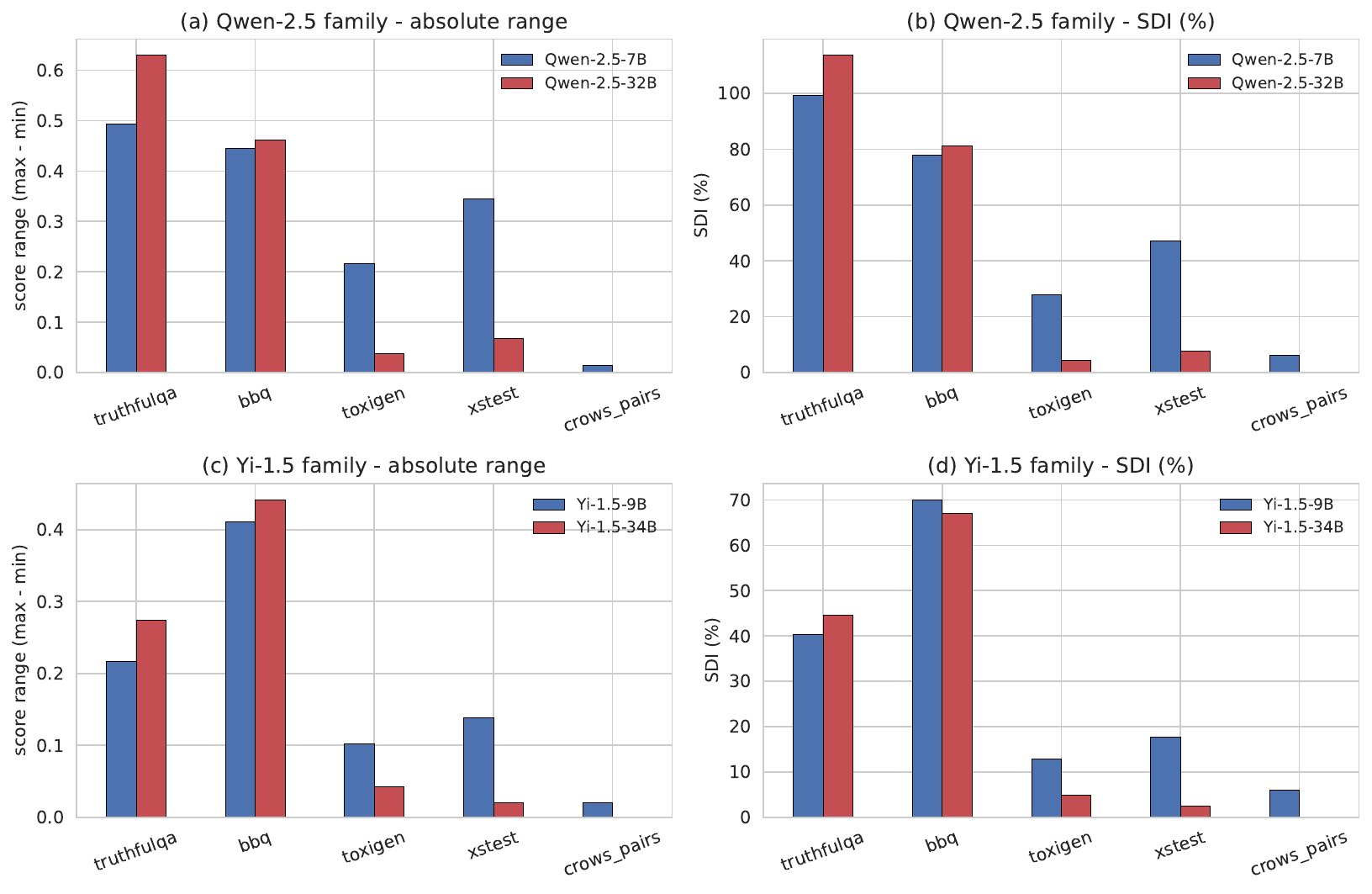}
\caption{Cross-family conservative-core absolute range and SDI:
top row Qwen-2.5 (7B vs 32B), bottom row Yi-1.5 (9B vs 34B), same
sub-envelope as Table~\ref{tab:scale}. The three-vs-two
collapse-vs-persist partition is consistent across both families.}
\label{fig:scale}
\end{figure}

\section{Per-package configuration table for the cross-package case study}
\label{appx:xharness-config}

We support the §\ref{sec:f-cross-harness} non-equivalence reading
with the per-package configuration that each runner actually
executes for the same nominal anchor (Qwen2.5-7B-Instruct, bf16,
greedy, 0-shot, 300 examples, seed 42). Items that differ
\emph{materially} across packages are marked $\bigtriangleup$;
items where the package's stated default matches the anchor are
marked $=$. Differences are not always the package's fault: for
some axes the package hard-codes a choice inside the task
definition that is not user-controllable without modifying the
installed task module. This is the operational content of
``implementation non-equivalence'' in §\ref{sec:f-cross-harness}.

\vspace{4pt}
{\footnotesize
\begin{tabular}{@{}lll@{~~}l@{~~}l@{}}
\toprule
axis & lm-evaluation-harness 0.4.5 & HELM-lite 0.5.5+ & Inspect AI 0.3.21 \\
\midrule
TruthfulQA prompt           & raw question + answers          & raw question + answers          & raw question + A/B/C/D options \\
TruthfulQA candidate set    & canonical-answer logprob (MC1)  & multiple\_choice\_joint (concat) & A/B/C/D letter logprob \\
TruthfulQA scorer           & argmax over canonical-answer LL & rank joint-target LL            & argmax over letter LL \\
TruthfulQA gold map         & dataset MC1 truth set           & dataset MC1 truth set           & separate letter $\to$ truth map \\
\midrule
BBQ prompt                  & not native (skip)               & subject=all, MC-joint adapter    & MC prompt via inspect\_evals/bbq \\
BBQ candidate set           & --                              & MC-joint concat                  & A/B/C choice logprob \\
BBQ scorer                  & --                              & rank joint-target LL            & inspect choice scorer \\
\midrule
ToxiGen prompt              & raw text + label cue            & not native (skip)                & not native (skip) \\
\midrule
chat template               & per-tokenizer apply\_chat\_template & package-internal jinja          & inspect-internal jinja \\
add\_generation\_prompt     & =True                           & package-internal               & package-internal \\
decoding (logprob axes)     & inert                           & inert                           & inert \\
decoding (free-form path)   & free-form not used here         & free-form not used here         & free-form not used here \\
few-shot                    & 0 ($=$ anchor)                  & 0 ($=$ anchor)                  & 0 ($=$ anchor) \\
chunking / max\_new\_tokens & N/A for logprob                 & N/A for logprob                 & N/A for logprob \\
\bottomrule
\end{tabular}
}

\vspace{4pt}
The 22-point TruthfulQA spread is therefore not separable into
``prompt'', ``parser'', ``scorer'', or ``chat-template'' contributions
without modifying the task definitions inside three different
installed packages, which is what we mean by \emph{implementation
non-equivalence between nominally similar tasks}. The minimum
operational reading of Table~\ref{tab:cross-harness} that survives
this caveat is the one stated in §\ref{sec:f-cross-harness}:
disclosing a configuration is not, on its own, sufficient to
reproduce a score across mainstream evaluation packages today.

\section{Adversarial-set SHAP attribution}
\label{appx:adversarial-shap}

\begin{figure}[H]
\centering
\includegraphics[width=\columnwidth]{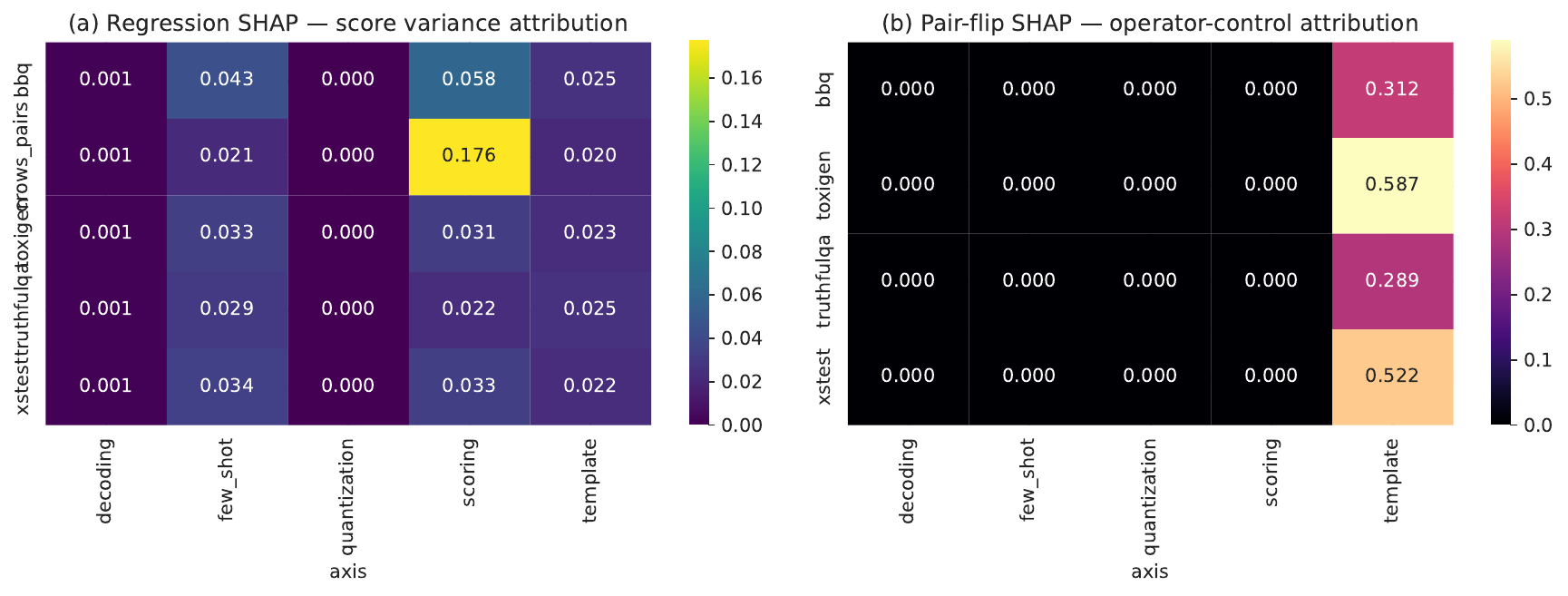}
\caption{Mean $|$SHAP$|$ per axis for (a) score regression and (b)
operator-controllable pair-flip classification. Higher = the axis
moves the model output (or the ranking outcome) more, on the
specified benchmark. The \texttt{quantization} column is shown
for completeness only: quantization is held constant (NF4) on
this adversarial grid and contributes $0.000$ in every cell by
construction.}
\label{fig:adversarial-shap}
\end{figure}

Figure~\ref{fig:adversarial-shap} shows mean $|$SHAP$|$ value per
axis for two surrogate models trained on the §\ref{sec:f-adversarial}
adversarial grid: (a) a regressor of score on $(model, benchmark,
\text{axes\,...})$, attributing variance, and (b) a per-pair binary
classifier of $A>B$ on axes alone (model identity excluded), attributing
ranking-control. Both are LightGBM with 300--400 trees,
\texttt{learning\_rate=0.05}, \texttt{max\_depth=4--6}, seed 42; the
surrogate is used only to localise the axis the operator exploits, not
to certify rank flips (those are exact counts on the observed grid).
Attribution at a different unit of analysis — attention consistency
at the token scale~\citep{lan-etal-2025-attention} — is complementary
to this configuration-axis-level attribution and orthogonal to the
identification question we close in Prop.~\ref{prop:nonid}.
The feature matrix contains
$(\text{model}, \text{benchmark}, c_{\text{template}},
c_{\text{decode}}, c_{\text{few-shot}}, c_{\text{score}})$ as
applicable per row; quantization is constant NF4 on this adversarial
grid, $c_{\text{quant}}$ is not a feature, and the 54-cell BF16
conservative-core sweep (App.~\ref{appx:bf16}) is excluded from
this surrogate's training data.

\section{Adversarial-set commit-stamped rules}
\label{sec:adversarial-rules}

The practice-derived envelope used in §\ref{sec:f-adversarial}
is fixed by the rules below. The full \texttt{rules.md} file
\textbf{(SHA-256 \texttt{11fc8d8b\dots1e524f5})} is committed to
the released artefact bundle and is hashed at the start of
\texttt{rank\_flip\_exact.py}. A configuration is included
iff every axis value is justified below; configurations outside
this envelope are excluded from the rank-flip enumeration. Each
axis value is tagged \emph{core} (cited alignment-benchmark or
harness default) or \emph{stress} (practice-adjacent but without a
fixed alignment-benchmark precedent on TruthfulQA / BBQ /
ToxiGen / CrowS-Pairs / XSTest).

\paragraph{Prompt template.} Templates T1 (minimal/direct,
App.~\ref{appx:templates}), T2 (role-framed system preamble), T3
(\texttt{lm-evaluation-harness}-style detailed instruction), T4
(chain-of-thought style). Tier and provenance:
\textbf{[core]} T1 matches the originals of TruthfulQA, BBQ,
ToxiGen and the HELM minimal-prompt path;
\textbf{[core]} T2 matches the role-framed variants studied by
\citet{sclar2024quantifying};
\textbf{[core]} T3 matches the canonical
\texttt{lm-evaluation-harness} few-shot MCQ format
\citep{biderman2024lessons};
\textbf{[stress]} T4 instantiates the chain-of-thought
prompting pattern \citep{wei2022chain,kojima2022large} with the
``Answer:'' suffix matching lm-evaluation-harness MCQ
post-processing \citep{biderman2024lessons} --- we do not have a
specific TruthfulQA / BBQ / ToxiGen / CrowS-Pairs / XSTest
benchmark paper that adopts T4-style CoT as a default, so T4 is
included as a stress-tier adjacent-literature axis rather than a
documented alignment-benchmark default.

\paragraph{Decoding.}
\textbf{[core]} \emph{greedy} ($T{=}0$, $p{=}1.0$,
\texttt{do\_sample=False}) is the alignment-benchmark default in
all three frameworks
\citep{biderman2024lessons,liang2023holistic,inspectai2024}.
The two sampling settings, \textbf{[core]} \emph{moderate}
($T{=}0.3$, $p{=}0.9$) and \textbf{[stress]} \emph{diverse}
($T{=}0.7$, $p{=}0.9$), are author-chosen practice-adjacent
stress levels. \citet{biderman2024lessons} motivate reporting
sampling hyperparameters and \citet{sclar2024quantifying} document
prompt/evaluation sensitivity to such choices, but these exact
values are not framework defaults. We do not claim either
sampling level is a published default; we include them to span
the practice-relevant low-to-moderate temperature range.

\paragraph{Few-shot count.} $\{0, 3, 5\}$. 0-shot is the
alignment-benchmark default (TruthfulQA \citep{lin2022truthfulqa},
BBQ \citep{parrish2022bbq}, ToxiGen \citep{hartvigsen2022toxigen},
XSTest \citep{rottger2024xstest} originals); 3-shot and 5-shot are
canonical few-shot counts in the lm-evaluation-harness MMLU/BBH
configurations \citep{biderman2024lessons} and the HELM-lite
default few-shot specification \citep{liang2023holistic}. Counts
$>5$ are out of envelope: rare in alignment-benchmark publications
and they shrink the adversarial set without precedent.

\paragraph{Scoring.} \texttt{free\_form} (generate text +
regex-extract label, used in HELM, original BBQ, original XSTest)
or \texttt{logprob} (score each candidate continuation, argmax,
used in lm-eval-harness MCQ tasks and Open LLM Leaderboard).
LLM-judge and embedding-similarity scorers are out of envelope
(different fidelity regime; would conflate model-vs-judge variance
with benchmark-vs-implementation variance).

\paragraph{Quantisation.} The 612-cell grid is NF4 throughout; BF16 was
run only on a 54-cell matched conservative-core sweep
(App.~\ref{appx:bf16}) and is not part of the adversarial
envelope. Precision is therefore not included as a free axis in
§\ref{sec:f-adversarial}; we do not infer any monotonicity of
$\rho_{\text{flip}}{=}\min(c_{A>B},c_{B>A})/N$ from the NF4-only
grid. Adding precision configurations changes both the counts and
the denominator, so the observed flip rate could increase or
decrease depending on which side those configurations support; the
54-cell BF16 conservative-core sweep (App.~\ref{appx:bf16}) directly
maps the precision axis on the conservative core rather than relying
on a monotonicity assumption.

\paragraph{Combination constraints.} Two combinations are excluded
as malformed practice:
(1) \texttt{logprob}~+~non-greedy decoding (logprob is deterministic
over candidates; sampling does not apply), and
(2) CrowS-Pairs~+~\texttt{free\_form} (benchmark only supports the
logprob/pair format). Both are already enforced by
\texttt{ExperimentConfig.is\_valid}.

\paragraph{Commit-stamp commitment.} Modifying the rules above
after seeing results would require explicit annotation in
§\ref{sec:threats}. We use the term ``commit-stamp'' rather than
``pre-registration'' because the rules are version-pinned to a
git commit recorded in the artifact (hash redacted for blind
review), not externally pre-registered with a registry.

%% --- BEGIN: moved from body to fit 8pp limit ---
\section{Cross-package case study: non-equivalence of package-default evaluation stacks}
\label{appx:cross-package}
\label{sec:f-cross-harness}

§\ref{sec:f-adversarial} sweeps configurations inside a single
in-house harness. As supporting context we re-run the same nominal
anchor configuration of one model on three benchmarks with native
coverage in at least one of three widely used evaluation packages
(the score-range column in Tab.~\ref{tab:cross-harness} is reported
only on rows where ${\geq}2$ packages cover the benchmark; ToxiGen
is only natively shipped by lm-evaluation-harness in our package
versions and contributes one in-package number, not a range).

\paragraph{Anchor.} Qwen2.5-7B-Instruct in bf16, greedy
($T{=}0$, $p{=}1.0$, \texttt{max\_new\_tokens}${=}96$), 0-shot,
$300$ examples, seed 42. Each package draws its 300 examples
through its native dataset adapter and seed-42 sampler; the
resulting item IDs are not enforced to be identical across the
three packages, and we treat residual item-set drift as an
additional confound on the cross-package range column rather than
an isolatable axis. Packages:
\texttt{lm-evaluation-harness} 0.4.5
\citep{biderman2024lessons,lmeval_v045},
HELM-lite 0.5.5+ \citep{liang2023holistic,helmlite_v055}, and
Inspect AI 0.3.21 \citep{inspectai2024,inspectai_v0321}
(\texttt{anchor\_configs.json} in the released bundle).

\paragraph{Coverage and non-equivalence.} Native task availability
varies (lm-eval ships TruthfulQA-MC1 + ToxiGen; HELM-lite ships
TruthfulQA + BBQ; Inspect ships TruthfulQA + BBQ); the score range
is reported only when ${\geq}2$ packages cover the benchmark.
Despite the shared nominal anchor, the three packages do not score
the same candidate object: lm-eval logprobs over the
\texttt{truthfulqa\_mc1} canonical-answer set, HELM uses the
\texttt{multiple\_choice\_joint} concat-then-rank adapter, Inspect
takes A/B/C/D letter logprobs with a separate ground-truth map (BBQ
analogously: HELM \texttt{multiple\_choice\_joint} vs Inspect
\texttt{choice} scorer). Chat-template and prompt boilerplate also
differ; full per-package configuration in
App.~\ref{appx:xharness-config}. Table~\ref{tab:cross-harness} is
therefore evidence of \emph{implementation non-equivalence} between
nominally identical benchmark--anchor pairs, not a per-axis
attribution.

\begin{table}[t]
\centering\small
\caption{Score per package at the same nominal anchor configuration
on one model. Scores are non-equivalent measurands (different
candidate sets and prompt boilerplate per package). The
``range'' column is reported with that caveat.}
\label{tab:cross-harness}
\begin{tabular}{@{}lrrrr@{}}
\toprule
benchmark & lm-eval & HELM & Inspect & range \\
\midrule
TruthfulQA & 0.460 & 0.617 & 0.677 & 0.217 \\
BBQ        & --    & 0.910 & 0.683 & 0.227 \\
ToxiGen    & 0.753 & --    & --    & --    \\
\bottomrule
\end{tabular}
\end{table}

\begin{table}[t]
\centering\footnotesize
\caption{What differs across the three package-default evaluation
stacks at the same nominal anchor (TruthfulQA row, illustrative).
$\bigtriangleup$~material difference, $=$~matches anchor. Full
matrix (all three benchmarks, all axes) in
App.~\ref{appx:xharness-config}.}
\label{tab:xharness-diffs}
\setlength{\tabcolsep}{2pt}
\scriptsize
\begin{tabular}{@{}llll@{}}
\toprule
axis & lm-eval & HELM-lite & Inspect AI \\
\midrule
candidate set & MC1 canon.\ $\triangle$ & MC-joint concat $\triangle$ & A/B/C/D letters $\triangle$ \\
scorer        & argmax canon.\ LL       & rank joint LL                & argmax letter LL \\
gold map      & MC1 truth set            & MC1 truth set                & sep.\ letter map \\
chat template & per-tokenizer            & pkg jinja $\triangle$         & inspect jinja $\triangle$ \\
few-shot      & 0 ($=$)                   & 0 ($=$)                       & 0 ($=$) \\
\bottomrule
\end{tabular}
\end{table}

\paragraph{What this case study supports.} A community user picking
between three popular packages and holding the nominal configuration
constant observes scores on the same model and benchmark that
differ by 22 points; the minimum reading of
Table~\ref{tab:cross-harness} that survives the non-equivalence
caveat is that \emph{disclosing a configuration is not, by itself,
sufficient to reproduce a score across mainstream evaluation
packages today}. Decomposing this gap into prompt-template,
chat-template, and parser components requires task-definition edits
inside the three installed packages and is left for future work
(§\ref{sec:threats}).

\section{Aggregate variance partition: SDI, CFR, $\rho$}
\label{appx:variance-partition}

The within-grid variance partition we recover from the
612-cell sweep is consistent with the headline rank-flip
result; the cross-package case study (\S\ref{sec:f-cross-harness})
provides separate illustrative context. Details summarised here
and detailed in Appendices~\ref{appx:absranges},
\ref{appx:anova} and \ref{appx:boot}.

\paragraph{Score dispersion (SDI).} Mean Score Dispersion
$(\max\!-\!\min)/\bar s$ across (model, benchmark) cells is
$54\%$; the maximum is $102.4\%$ on Qwen-2.5-7B / TruthfulQA, with
max-statistic configuration-bootstrap sensitivity range $[76.1, 109.1]\%$. The absolute swing
on that cell is $55$ percentage points
(s$_{\min}{=}0.140$ to s$_{\max}{=}0.693$; Appendix~\ref{appx:absranges}).
Only CrowS-Pairs --- which admits only the logprob path --- stays
below $35\%$ SDI on all three models (App.~\ref{appx:extra-figs}
and~\ref{appx:boot}).

\paragraph{Pass/fail (CFR).} At the illustrative threshold
$\theta{=}0.5$, $\CFR_\theta$ peaks at $51.1\%$ on Mistral-7B /
TruthfulQA (the grid is split $24/24$); at $\theta{=}0.7$,
$\CFR_\theta$ peaks at $49.6\%$ on Qwen-2.5-7B / BBQ. What an
auditor calls a ``passing'' score is, on these cells, an
implementation-grid balance issue, not a stable property of
(model, benchmark). Full per-cell tables are in
Appendix~\ref{appx:cfr-tables}.

\paragraph{Variance shares ($\rho$).} Four-way Type-II ANOVA on the
free-form slice (factors $\{$model, template, decoding,
few-shot$\}$) gives implementation-to-model variance ratios
$\rho$=3.8 (BBQ), \textbf{14.9 [3.2, 249.2]} (ToxiGen), 0.7
(TruthfulQA), 2.4 (XSTest), where the bracketed interval is the
within-grid bootstrap range on the ToxiGen headline
(App.~\ref{appx:multiplicity}). The directional reading ---
implementation main effects exceed model main effects on three of
four benchmarks, with only TruthfulQA model-dominated --- is robust;
the magnitude on ToxiGen is not. The 14.9 point estimate should be
read as evidence of strong amplification, not as a precisely known
multiplier: the interval is wide because the model-variance share
is small ($1.8\%$, so the ratio is unstable near zero) and
interaction shares are large (41--54\%). We therefore use $\rho$ as
an order-of-magnitude qualitative ordering rather than an
identified estimate. Type-III robustness (XSTest drops to
$\rho{=}1.3$), per-benchmark bootstrap intervals, and a parse-rate
${\geq}0.95$ validated-subset re-run that preserves the qualitative
ordering are reported in Appendix~\ref{appx:typeIII} and
Appendix~\ref{appx:parse}. See Table~\ref{tab:mvi}.

\paragraph{Scoring path as a first-class axis.} Switching only the
scoring pipeline (free-form regex parse $\leftrightarrow$ logprob
argmax) moves the score by up to $0.468$ on
Qwen-2.5-7B / TruthfulQA (per-cell numbers in
App.~\ref{appx:extra-figs}, Fig.~\ref{fig:scoring}); in MTMM terms
\citep{campbell1959mtmm} the two paths score non-equivalent
measurands and we report the gap rather than mark either as
canonical.

\begin{table}[t]
\centering\small
\caption{Type-II ANOVA on the free-form slice: raw-$\eta^2$ shares (\%)
and impl/model ratio $\rho$. CrowS-Pairs excluded (free-form slice
empty). Per-cell SDI table is in App.~\ref{appx:sdi-extra}
(Tab.~\ref{tab:sdi}); the variance-decomposition figure is
Fig.~\ref{fig:anova} in App.~\ref{appx:extra-figs}.}
\label{tab:mvi}
\begin{tabular}{lrrrr}
\toprule
Benchmark & Model & Impl. & Inter. & $\rho$ \\
\midrule
BBQ        &  9.1 & 34.7 & 44.8 &  3.8 \\
ToxiGen    &  1.8 & 26.6 & 48.4 & 14.9 \\
TruthfulQA & 25.4 & 18.7 & 41.0 &  0.7 \\
XSTest     &  9.3 & 22.2 & 53.9 &  2.4 \\
\bottomrule
\end{tabular}
\end{table}

\paragraph{Exploratory note: cross-family scale probe
(Qwen-2.5 7B$\to$32B + Yi-1.5 9B$\to$34B).} Re-running the
conservative-core sub-envelope at the larger sibling in two
families (App.~\ref{appx:scale-probe}, Tab.~\ref{tab:scale})
narrows the conservative-core score range to near-zero on ToxiGen,
XSTest, and CrowS-Pairs in \emph{both} families, while leaving it
large on TruthfulQA and BBQ in both (Qwen-2.5: $0.494{\to}0.631$
and $0.445{\to}0.462$; Yi-1.5: $0.217{\to}0.275$ and
$0.411{\to}0.442$). We report this as a \emph{bounded two-family
observation on the conservative-core sub-envelope}, not a scaling
law: it reduces (but does not rule out) the single-family confound on this 4-config
envelope, but two families is not evidence of a benchmark
taxonomy-by-scale, and we make no claim about Mistral, Llama, or
closed-weight models.

\section{Disclosure template (GRID card) commentary}
\label{sec:protocol}

Given §\ref{sec:f-adversarial} and §\ref{sec:f-cross-harness}, a
benchmark score is reproducible only when accompanied by both its
configuration and its harness. The \textbf{GRID card}
(App.~\ref{appx:gridcard}) is a compact research-level disclosure
whose fields cover the implementation axes plus the harness
identifier and a minimum statistical summary. It is a starter
artefact: no adopter, enforcement, or compliance criteria is
specified, and it does not resolve the \emph{standing} problem
(a claimant who picks the configuration neighbourhood can game it,
as §\ref{sec:f-adversarial} shows). Turning configuration
disclosure into a deployable verification instrument is
institutional work this paper does not do.

%% --- END moved-from-body ---

%% --- Item-subset robustness, moved to fit 8pp ---
\section{Item-subset robustness on TruthfulQA conservative core}
\label{appx:item-subset}

\paragraph{Robustness to item-subset choice
(3 stratified 80\% subsamples).}
The headline is conditioned on a single fixed item subset
($n{=}295$ on TruthfulQA, seed 42). To check whether the
pairwise-disagreement metric is itself subset-idiosyncratic, we
re-ran the 12-config conservative-core slice on three independent
stratified 80\% subsamples ($n{\approx}236$ items each, seeds 43--45)
on TruthfulQA, yielding 12 (model, configuration) cells $\times$ 3
subsamples = 36 raw-score measurements
(Tab.~\ref{tab:tqa-resample}). The per-cell raw-score range across
the three subsamples is small: median $1.49$~pp (mean $1.63$~pp,
max $3.83$~pp on Yi/T1/free-form). The within-cell nearest-pair
gap is large enough on T1/free-form (every seed $\geq 15$~pp) that
no plausible $\sim$2--4~pp resample swing can flip an ordering;
for the other three conservative-core configurations the
nearest-pair gaps are smaller (mostly 2--8~pp, with one
T3/free-form/seed-44 cell where Qwen and Yi land at exactly
$0.6255$ each --- a tie broken by a deterministic alphabetical
convention rather than a strict raw-score ordering). The
config-level orderings are nevertheless stable: on each of the
four conservative-core configurations the induced (Qwen, Mistral,
Yi) ordering is the same on every subsample (0/12 cells flip
under the deterministic tie convention; the one tied cell is
flagged in Tab.~\ref{tab:tqa-resample}). $\rho_{\text{flip}}$ at
$|C|{=}4$ is discretised in $0.25$ increments, so its trivial
cross-subsample stability (constant per-pair value: 0.250 /
0.250 / 0.500 for qwen-mistral / qwen-yi / mistral-yi on every
subsample) carries less information than the raw-score ranges;
the latter is the substantive evidence. We treat this as a
conservative-core smoke test on TruthfulQA: no instability was
observed across these three 80\% subsamples (modulo one Qwen/Yi
tied cell that does not contribute a strict ordering on its own),
but full-envelope ($|C|{=}48$) item resampling and resampling on
the other four benchmarks remain future work; we therefore do not
generalise this stability beyond the 12-cell conservative core on
TruthfulQA.

\begin{table}[t]
\centering\small
\caption{TruthfulQA conservative-core ($|C|{=}4$) item-subset
robustness on 3 stratified 80\% subsamples (rs43--rs45,
$n{\approx}236$). Per-cell raw-score range across subsamples (pp)
plus per-config ordering on (Qwen, Mistral, Yi). All four
orderings are invariant across all three subsamples.}
\label{tab:tqa-resample}
\setlength{\tabcolsep}{3pt}
\footnotesize
\begin{tabular}{@{}llcccc@{}}
\toprule
config & ordering (rs43--45) & qwen $\Delta$ & mistral $\Delta$ & yi $\Delta$ & cell-flips \\
\midrule
T1/free-form & yi $>$ mistral $>$ qwen & 1.70 & 1.28 & 3.83 & 0/3 \\
T1/logprob   & mistral $>$ qwen $>$ yi & 0.43 & 2.55 & 0.43 & 0/3 \\
T3/free-form & qwen $\geq$ yi $>$ mistral$^\dagger$ & 1.71 & 1.28 & 2.98 & 0/3$^\dagger$ \\
T3/logprob   & mistral $>$ qwen $>$ yi & 0.00 & 2.13 & 1.28 & 0/3 \\
\midrule
\multicolumn{2}{@{}l}{12 cells aggregated} & \multicolumn{4}{c}{median 1.49~pp, mean 1.63~pp, max 3.83~pp} \\
\bottomrule
\end{tabular}
\\
\footnotesize $^\dagger$ T3/free-form, seed 44: Qwen and Yi
land at $0.6255$ each (exact tie). Listed as
qwen $\geq$ yi $>$ mistral under the deterministic alphabetical
tie-break convention used by \texttt{rank\_flip\_stratified.py};
the underlying ordering is a tie, not a strict reversal.
\end{table}

%% --- Conservative-core saturation, moved to appendix to fit 8pp ---
\section{Conservative-core sub-envelope saturation}
\label{appx:envelope-choice}

\paragraph{Robustness to envelope choice.} A natural objection
to §\ref{sec:f-adversarial} is that the envelope itself is
hand-curated and could be argued to maximise volatility. We
re-compute the same exact rank-flip metric on a \emph{conservative
core} sub-envelope that drops every contestable axis value: only
templates T1 (minimal) and T3 (canonical lm-eval-harness MCQ);
greedy decoding only; 0-shot only; both scoring paths; NF4 (4
configs per free-form benchmark, 2 for CrowS-Pairs). This is the
narrowest audited conservative-core sub-envelope inside the
core tier --- a smoke test, not a representative sample of
practice. Table~\ref{tab:adv-robustness} reports the side-by-side
metrics.

\begin{table}[t]
\centering\small
\caption{Rank-flip robustness on a conservative-core sub-envelope
($|C|{=}4$ free-form, $|C|{=}2$ CrowS). Same metric as
Table~\ref{tab:adversarial}; recall $\rho_{\text{flip}}^{\max}$ is
bounded above by $0.5$ by combinatorial construction at $|C|{=}4$
so the ``four of five attain $0.5$'' result is read as ``these
benchmarks attain the maximum possible value at this envelope
size'', not as a generic 50\% failure rate.}
\label{tab:adv-robustness}
\begin{tabular}{lrrrr}
\toprule
 & \multicolumn{2}{c}{full envelope} & \multicolumn{2}{c}{conservative core} \\
\cmidrule(lr){2-3}\cmidrule(lr){4-5}
benchmark & $|C|$ & $\rho_{\text{flip}}^{\max}$ & $|C|$ & $\rho_{\text{flip}}^{\max}$ \\
\midrule
TruthfulQA  & 48 & 0.479 & 4 & \textbf{0.500} \\
BBQ         & 48 & 0.438 & 4 & \textbf{0.500} \\
ToxiGen     & 48 & 0.354 & 4 & \textbf{0.500} \\
XSTest      & 48 & 0.292 & 4 & \textbf{0.500} \\
CrowS-Pairs & 12 & 0.250 & 2 & 0.000 \\
\bottomrule
\end{tabular}
\end{table}

The conservative-core check therefore evidences \emph{existence}
of operator-controllable pairwise reversals even under a strict
4-config sub-envelope, but at $|C|{=}4$ the metric is
upper-bounded by $0.5$ by combinatorial construction (any 2--2
split already saturates the bound); the slice cannot estimate the
\emph{magnitude} of envelope-driven reversal --- only that
reversals exist on the narrowest audited sub-envelope. Four of
five benchmarks reach this ceiling, meaning that on each of those
benchmarks the maximising model pair splits the four
conservative-core configurations 2--2. The ceiling is attained for
at least one model pair per benchmark, not necessarily for every
pair or every cell --- even the narrowest audited conservative-core
sub-envelope inside the core tier contains an $A{>}B$ and a $B{>}A$
configuration in equal number for the maximising pair.
CrowS-Pairs collapses to $0.0$ at $|C|{=}2$ (both configs agree).
The full-envelope ``$5$--$6$ distinct orderings'' headline does
depend on the broader envelope (T2/T4, few-shot ${>}0$); the
qualitative existence of pairwise reversals on the four free-form
benchmarks is preserved on the conservative-core sub-envelope, but
the $47.9\%$ TruthfulQA $\rho_{\text{flip}}^{\max}$ magnitude is a
full-envelope quantity and should not be inferred from the
saturated $|C|{=}4$ core (where $\rho_{\text{flip}}$ takes only
the values $0,0.25,0.5$).

%% --- Claim-scope table, moved to appendix to fit 8pp ---
\section{Claim-scope table}
\label{appx:claim-scope}

\begin{table}[H]
\centering\footnotesize
\caption{Claim scope. Headline numbers in §\ref{sec:f-adversarial} sit
in the \textbf{Supported} column; §\ref{sec:f-cross-harness} sits in
\textbf{Suggestive}; everything else is \textbf{Out of scope}.}
\label{tab:claim-scope}
\setlength{\tabcolsep}{3pt}
\begin{tabular}{@{}p{0.31\columnwidth}p{0.31\columnwidth}p{0.31\columnwidth}@{}}
\toprule
\textbf{Supported} & \textbf{Suggestive} & \textbf{Out of scope} \\
\midrule
Configuration-conditional pairwise-disagreement on the 612-cell
NF4 grid (per-benchmark rank-flip envelope: full mixed
$|C|{=}48$, free-form-only $|C|{=}36$, logprob-only $|C|{=}12$;
CrowS-Pairs $|C|{=}12$ logprob-only),
3 models $\times$ 5 benchmarks, fixed item subset.
&
Cross-package score spread at one anchor (illustrative; packages
score non-equivalent objects).
&
Population-level rank stability across model families or model
sizes; closed-weight models; jailbreak / multiturn benchmarks. \\
\addlinespace[2pt]
That an \emph{evaluator inside this envelope} can move pairwise
verdicts on up to 47.9\% of the $|C|{=}48$ TruthfulQA harness
configurations and produce all 6 orderings on XSTest under the full
practice-derived envelope (Tab.~\ref{tab:adversarial}); on the
core-tier envelope (drops T4 and diverse decoding) TruthfulQA
$\rho_{\text{flip}}^{\max}$ is 40.7\% on $|C|{=}27$ and XSTest
admits 5/6 orderings (Tab.~\ref{tab:adv-envelope-tier}).
&
Cross-family scale-probe collapse on conservative-core
(Qwen 7$\to$32B, Yi 9$\to$34B; Tab.~\ref{tab:scale}).
&
A scaling law; a per-axis decomposition of cross-package spread;
an enforceable governance instrument. \\
\addlinespace[2pt]
Variance partition direction (impl.\ main effects $\geq$ model
main effects on three of four free-form benchmarks).
&
Type-III re-analysis robustness (XSTest is boundary-sensitive).
&
A precise $\rho$ multiplier (interaction shares 41--54\% make
ratios qualitative). \\
\addlinespace[2pt]
GRID card as a research-level disclosure template.
&
NF4-vs-BF16 score-range comparability on the conservative-core sub-envelope (54
cells; Qwen + Mistral comparable, Yi picks up extra spread on
ToxiGen and XSTest).
&
GRID card as a regulator-ready compliance instrument; a full
NF4-vs-BF16 sweep on the full 612-cell grid. \\
\bottomrule
\end{tabular}
\end{table}

\end{document}